\documentclass{article}

\usepackage[main,final,nonatbib]{neurips_2026}

\usepackage[T1]{fontenc}    
\usepackage{hyperref}       
\usepackage{url}            
\usepackage{booktabs}       
\usepackage{amsmath}       
\usepackage{amsfonts}       

\usepackage{nicefrac}       
\usepackage{microtype}      
\usepackage[table]{xcolor}         
\usepackage{graphicx}      
\usepackage{amssymb}       
\usepackage{booktabs}
\usepackage{multirow}
\usepackage{makecell}
\usepackage{subcaption}

\usepackage{algorithm}
\usepackage{algpseudocode}

\usepackage{longtable}
\usepackage{array}
\usepackage{siunitx}

\newcommand{\se}[1]{%
  {\fontsize{5.5}{6.5}\selectfont\textpm\,#1}%
}

\usepackage[normalem]{ulem}
\newcommand{\modelname}{ActionUNet}
\definecolor{darkgreen}{rgb}{0.0, 0.5, 0.0}

\title{ActionUNet: Improving Robustness of VLA Models with Efficient Multi-scale Fine-tuning}

\author{%
  Di Zhu\thanks{Equal contribution.}
  \qquad
  Ziheng Yan\footnotemark[1]
  \qquad
  Fang Wan\thanks{Corresponding author.}
  \\
  University of Chinese Academy of Sciences
  \\
  \texttt{\{zhudi25,yanziheng21\}@mails.ucas.ac.cn}
  \\
  \texttt{wanfang@ucas.ac.cn}
}

\begin{document}

\maketitle

\begin{abstract}
Vision-Language-Action (VLA) models have shown great promise for robotic manipulation by mapping multi-modal semantics to physical actions.
However, this mapping inherently struggles to align these coarse-grained semantics with fine-grained temporal execution. It leaves VLA models with limited generalization and insufficient robustness in cluttered environments.
To overcome this issue, we propose ActionUNet, an efficient multi-scale fine-tuning framework that enhances pre-trained VLA models with minimal computational cost. 
ActionUNet first constructs a lightweight 
temporal
U-Net within the temporal-aligned action feature space to fuse hierarchical structural priors, effectively bridging the scale gap between semantics and temporal executions. 
Recognizing that multi-scale modeling can disrupt microscopic temporal continuity and cause mechanical oscillations, ActionUNet then employs a conditional SIREN as a continuous action decoder.
Equipped with explicit second-order smoothness constraints, this decoder 
guarantees 
temporal continuity and 
reduces
high-frequency motion jitter. By smoothing temporal discontinuities from multi-scale fusion, this continuous formulation 
reduces
mechanical execution failures while preserving the base VLA model's generalization and manipulation robustness.
Extensive experiments on RoboTwin 2.0 and LIBERO-Plus benchmarks, together with real-world hard evaluations, demonstrate that ActionUNet significantly improves $\pi_{0.5}$ success rates by absolute 9.8\%, 6.1\%, and 11.4\%, respectively, while also generalizing to the regression-based OpenVLA-OFT backbone, highlighting its effectiveness and efficiency as a fine-tuning strategy.
Code and implementation details are available at
\url{https://github.com/Di-Zhu123/ActionUNet}.

\end{abstract}

\section{Introduction}

Vision-Language-Action (VLA) models, which combine language instructions with visual environmental perception, have emerged as a dominant paradigm for robotic manipulation tasks. With the rapid progress of vision-language models, a series of VLA approaches~\cite{RT2, OpenVLA, gr00t_n1, pi05} have demonstrated promising performance. 
However, existing VLA models still face critical challenges, including limited generalization ability~\cite{GeneralVLA} and insufficient robustness in cluttered or distracting environments~\cite{DistractedRobot}.

Current VLA models~\cite{RDT, OpenVLA-OFT, pi05} typically attach an action expert to the large vision-language model to predict future action sequences, prompted by a single language instruction. This action expert learns to translate the VLM's multi-modal representations directly into action trajectories, typically through a transformer decoder or a diffusion-based generative head. While VLMs excel at high-level task reasoning, they often struggle with precise, fine-grained action prediction~\cite{yuan2024robopoint,zhao2025manipbench}.

\begin{figure}
    \centering
    \includegraphics[width=1.0\linewidth]{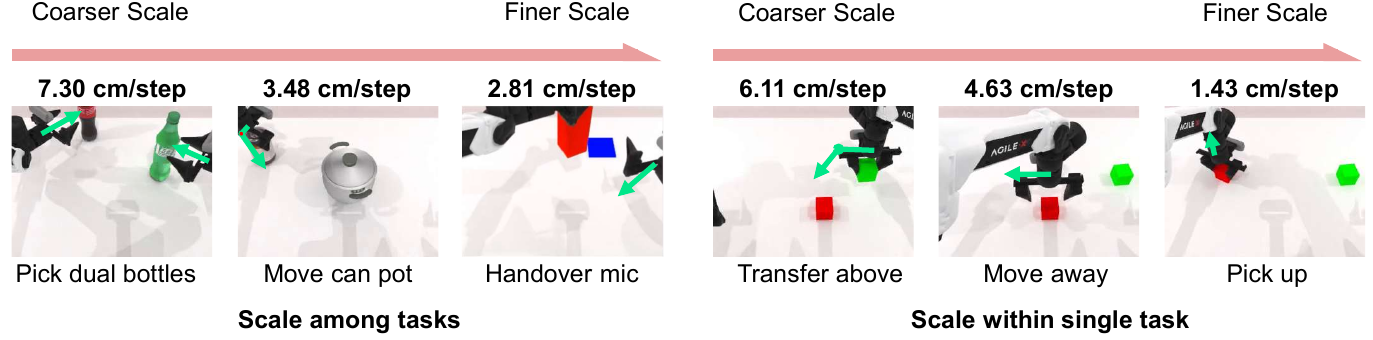}
    \caption{Illustration of multi-scale temporal dynamics. Per-step end-effector displacements vary significantly across different tasks (left) and over time within a single task ``Stack blocks two'' (right).}
    \label{fig:motivation}
    \vspace{-3mm}
\end{figure}

Central to robotic manipulation, action prediction must bridge high-level intentional inference for task planning with fine-grained, low-level control for execution. Furthermore, we observe that the per-step end-effector displacement varies significantly not only across different tasks (Fig.~\ref{fig:motivation} left) but also 
across different stages within the same task
(Fig.~\ref{fig:motivation} right). 
We characterize execution scale using the per-step end-effector displacement, where larger displacements correspond to coarser-scale motions and smaller displacements correspond to finer-scale motions (see Appendix~C).
This variance in per-step displacement indicates the differences in the temporal scale of actions, implying that VLA models are inherently required to possess multi-scale temporal representation capabilities. 
Consequently, directly mapping the coarse-grained multi-modal semantics to fine-grained action sequences suffers from a severe \textit{temporal scale mismatch} issue. This mismatch undermines the VLA models' robustness to visual observation noise, as the model over-relies on abstract semantics while failing to capture the multi-scale, fine-grained temporal cues necessary for resisting local perturbations.

To address this issue, 
we introduce multi-scale action modeling in the action feature space.
However, implementing such multi-scale representations risks breaking the temporal continuity of actions across different scales. This occurs because aggregating actions from varying granularities often projects disjointed structural priors onto the continuous action space, causing interpolation errors at scale boundaries. 
Consequently, the generated trajectories often exhibit undesired non-smooth transitions~\cite{liu2026learning}, leading to mechanical oscillations and further reducing the execution robustness of VLA models.
Therefore, a core challenge lies in equipping VLA models with multi-scale structural priors while simultaneously enforcing action smoothness constraints. Addressing this through efficient fine-tuning strategies has become an urgent necessity.

In this paper, we propose ActionUNet, an efficient multi-scale fine-tuning framework that significantly enhances the robustness of pre-trained VLA models with very low training and inference costs. 
Specifically, ActionUNet first constructs a lightweight temporal U-Net architecture within the temporal-aligned action feature space, injecting explicit multi-scale priors into the action generation process through hierarchical feature encoding and decoding. 
Then, ActionUNet introduces a conditional SIREN as a continuous action decoder to efficiently leverage the pre-trained VLA's action features. This architecture directly incorporates these action features as conditional inputs to preserve intrinsic temporal alignment, while simultaneously mitigating high-frequency motion jitter via explicit second-order smoothness constraints to 
guarantee 
temporal continuity.

By integrating the hierarchical structural priors from the temporal U-Net with the continuous temporal constraints of the conditional SIREN, ActionUNet effectively bridges the scale gap between coarse-grained multi-modal semantics and fine-grained physical execution. This ensures the generation of smooth, jitter-resistant action trajectories to enhance the model's resilience against environmental perturbations, while fully leveraging the pre-trained VLA to improve fine-tuning efficiency.

Our main contributions are summarized as follows:

\begin{itemize}
    \item We propose ActionUNet, a simple-but-effective multi-scale fine-tuning framework that constructs a 
    temporal U-Net to endow pre-trained VLA models with hierarchical structural priors, effectively mitigating the scale mismatch problem at a minimal computational cost.
    
    \item We design a conditional SIREN decoder for continuous action generation, which efficiently leverages the multi-scale fused action features as conditional inputs. This design preserves intrinsic temporal alignment while imposing explicit second-order smoothness constraints to guarantee temporal continuity and mitigate high-frequency motion jitter.
    

    \item Extensive experiments on RoboTwin 2.0, LIBERO-Plus, and real-world dual-arm manipulation with hard setting demonstrate that ActionUNet significantly improves $\pi_{0.5}$ success rates by 9.8\%, 6.1\%, and 11.4\% in hard/perturbed settings, respectively. Its consistent gains on OpenVLA-OFT further highlight its effectiveness and compatibility as an efficient fine-tuning strategy.

\end{itemize}

\section{Related Work}

\textbf{Vision-Language-Action Model.}
Vision-Language-Action (VLA) models have emerged as a promising paradigm for robotic manipulation, typically building upon pre-trained Vision-Language Models (VLMs)~\cite{Paligemma, LLaVA, Qwen2-vl} to map visual and linguistic inputs to actions. Existing VLA architectures can be broadly categorized into two families: autoregressive models that discretize continuous actions into tokens~\cite{RT1, RT2, OpenVLA, pi0fast}, and diffusion-based or flow-matching-based models that generate continuous action chunks \cite{pi0, gr00t_n1, CogACT}. Despite these advances, robustness remains a key concern; prior work has shown that VLA models are vulnerable to visual corruptions \cite{vlatest, gao2026selfevolving}, and existing robustification strategies often rely on external large models or focus solely on visual perturbations \cite{BYOVLA, zhang2025gevrm}. Meanwhile, most VLA models adopt a single-scale action representation head and generate action sequences without explicit multi-scale or smoothness priors.

\textbf{Multi-scale Visuomotor Policy Learning.}
Robotic manipulation requires modeling actions at multiple temporal scales, since a policy must simultaneously capture long-horizon task progression and fine-grained local control.
Existing visuomotor approaches can be broadly grouped into two categories.
The first category learns discrete multi-scale representations of actions, often through vector-quantized tokenization, and performs hierarchical or scale-wise prediction~\cite{Carp, sheebaelhamd2025qfat}.
While such discretization can improve long-horizon reasoning and autoregressive efficiency, it introduces an information bottleneck that may hinder the transfer of rich visual features to downstream action prediction~\cite{The_Compression_Gap}.
The second category operates in continuous action space and introduces multi-scale structure either within the action generation process, by predicting action chunks at multiple resolutions~\cite{yashima2026hiflow} or decomposing actions into frequency components~\cite{zhang2026hipolicy}, or through multi-scale perception and hierarchical policy architectures that fuse features across sensor modalities~\cite{su2025densepolicy, H3DP, xue2025rdp}.
These methods, together with VLA-based multi-scale prediction that requires full-model retraining~\cite{huang2026mint}, demonstrate that coordinating global planning with local refinement is crucial, but their specially designed policy heads or training paradigms limit direct reuse of strong pre-trained VLA priors.
In contrast, we inject multi-scale structure directly into the action feature space, thereby bypassing the discretization bottleneck and the need for architecturally specialized heads, while retaining the base model's generalization with low fine-tuning overhead.

\section{Method}
\label{sec:method-multiscale}

\subsection{Preliminary}

In standard VLA fine-tuning, the action head inherits its weights from pre-training and decodes each timestep independently from high-level multi-modal features into a continuous action.
However, this single-scale mapping often fails to bridge the gap between coarse-grained multi-modal semantics and the multi-scale, fine-grained requirements of physical execution, leading to the scale mismatch issue. 

Our method, referred to as \textbf{ActionUNet}, addresses this issue by introducing two key modules: a temporal U-Net in the temporal-aligned action feature space that restores multi-scale structure, and a continuous action decoder that imposes explicit smoothness constraints. The temporal U-Net enriches the backbone action features with hierarchical priors, jointly capturing coarse task semantics and fine local control cues. These refined features are then decoded into a continuous-time trajectory via local SIREN fields and Mat\'ern-weighted aggregation, producing actions that are both spatially precise and temporally smooth.
By constructing the multi-scale prior directly in the continuous feature space, ActionUNet bridges the scale gap between coarse-grained multi-modal semantics and fine-grained actions, helps recover the multi-scale temporal cues needed for robust manipulation, and guarantees temporal continuity.

\begin{figure}
    \centering
    \includegraphics[width=1.0\linewidth]{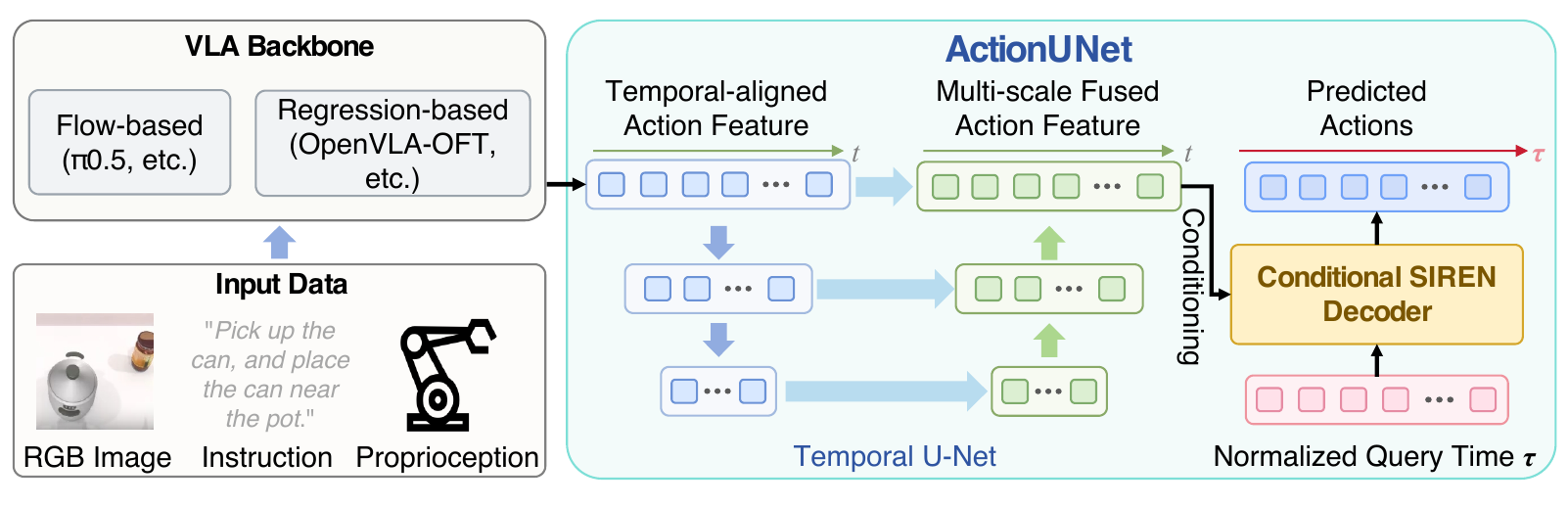}
    \caption{
Overview of ActionUNet. Given input observations and instructions, the VLA backbone first produces temporal-aligned action features. ActionUNet refines these features in the action feature space through multi-scale temporal mapping, and then decodes the fused features with a conditional SIREN into smooth continuous action trajectories.
    }
    \vspace{-3mm}
    \label{fig:method}
\end{figure}

\subsection{Temporal U-Net in Action Feature Space}
To address the scale mismatch and recover the fine-grained dynamics lacking in standard fine-tuning approaches, ActionUNet constructs a lightweight temporal U-Net that explicitly encodes multi-scale priors in the temporal-aligned action feature space. Let the action feature be denoted as $\mathbf{h}^{(0)} \in \mathbb{R}^{H \times d}$, where $H$ is the action horizon and $d$ is the feature dimension. The temporal U-Net progressively encodes the action feature into lower-resolution temporal representations and subsequently decodes it back to the original resolution. Skip connections recover fine-grained local details during reconstruction. The whole process is represented as:
\begin{align}
\mathbf{h}^{(\ell+1)} &= \mathrm{LN}\!\Big(\mathrm{GeLU}\!\big(\mathrm{Conv1D}^{(\ell)}(\mathbf{h}^{(\ell)})\big)\Big), \quad \ell = 0,\dots,L-1, \label{eq:unet_encoder}  \\
\tilde{\mathbf{h}}^{(\ell)} &= \mathrm{Fuse}^{(\ell)}\!\Big(\mathrm{Up}\!\big(\tilde{\mathbf{h}}^{(\ell+1)}\big),\; \mathbf{h}^{(\ell)}\Big), \quad \ell = L-1,\dots,0. \label{eq:unet_upsample} 
\end{align}
In Eq.~\eqref{eq:unet_encoder}, $\mathrm{Conv1D}^{(\ell)}$ denotes a one-dimensional convolution with a downsampling rate $r$ that halves the temporal resolution at each stage. The output is passed through a GeLU activation followed by Layer Normalization. Starting from $\mathbf{h}^{(0)} \in \mathbb{R}^{H \times d}$, after $L$ stages the hidden state $\mathbf{h}^{(L)}$ resides in $\mathbb{R}^{\lfloor H/r^L \rfloor \times d}$. 
{A fully connected layer transforms $\mathbf{h}^{(L)}$ into $\tilde{\mathbf{h}}^{(L)}$.}
Eq.~\eqref{eq:unet_upsample} restores temporal resolution in a coarse-to-fine manner. At each stage $\ell$, the current hidden state is first upsampled by a factor of two using nearest-neighbor interpolation ($\mathrm{Up}(\cdot)$). The upsampled feature is then concatenated with the corresponding encoder skip feature $\mathbf{h}^{(\ell)}$ and processed by a fusion block $\mathrm{Fuse}^{(\ell)}$, which consists of a single linear layer. The final output $\tilde{\mathbf{h}}^{(0)} \in \mathbb{R}^{H \times d}$ serves as the refined action feature.

\subsection{Conditional SIREN Decoder for Continuous Action Generation}

{Rather than directly mapping each refined feature vector $\tilde{\mathbf{h}}^{(0)}$ to an isolated action step, we interpret the action chunk as samples from a temporal-continuous trajectory over a normalized time $\tau \in [-1,1]$. To achieve this goal, we propose the conditional SIREN decoder, where the decoder produces a smooth function that can represent either the action trajectory itself (regression mode) or the flow-matching vector field (flow mode), depending on the training objective.}
This continuous formulation is motivated by standard robot trajectory-generation practice, where the robot's actions are often represented as smooth twice-differentiable or piecewise $C^2$ continuous curves to ensure continuity~\cite{kim2026time}.

\textbf{Local SIREN Fields.}
Let $\tilde{\mathbf{h}}^{(0)}_1,\dots,\tilde{\mathbf{h}}^{(0)}_H \in \mathbb{R}^{d}$ be the refined action features from the temporal U-Net, each associated with a uniformly spaced normalized time $\tau_1,\dots,\tau_H \in [-1,1]$.
For $i$-th feature $\tilde{\mathbf{h}}^{(0)}_i$, we construct a local implicit field $\Phi_i(\tau)$ that continuously maps the query time $\tau$ to the action space, conditioned on $\tilde{\mathbf{h}}^{(0)}_i$.
To endow every layer with explicit temporal awareness, we replace the standard static biases of SIREN~\cite{SIREN} with time-dependent biases defined as simple affine functions of $\tau$.
Formally, each hidden layer is a mapping
\begin{equation}
    \phi_m^{\tau} \colon \mathbb{R}^d \to \mathbb{R}^d, \quad
    \phi_m^{\tau}(\mathbf{z}) = \sin\!\bigl( \mathbf{W}_m \mathbf{z} + \mathbf{U}_m \tau + \mathbf{b}_m \bigr), \quad m = 0,\dots,M-1,
\end{equation}
where $\mathbf{W}_m \in \mathbb{R}^{d\times d}$, $\mathbf{U}_m \in \mathbb{R}^{d\times 1}$, and $\mathbf{b}_m \in \mathbb{R}^{d}$ are learnable parameters.
Starting from the refined feature $ \tilde{\mathbf{h}}^{(0)}_i$, the local SIREN field is the composition of these layers followed by a linear readout:
\begin{equation}
    \Phi_i(\tau) \;=\; \mathbf{W}_{M}\, (\phi_{M-1}^{\tau} \circ \cdots \circ \phi_0^{\tau})(\tilde{\mathbf{h}}^{(0)}_i) \;+\; \mathbf{b}_{M},
    \label{eq:siren-bias-time}
\end{equation}
with $\mathbf{W}_{M} \in \mathbb{R}^D\times d$, $\mathbf{b}_{M} \in \mathbb{R}^{D}$ where $D$ is the action dimension.
Because every $\phi_m^{\tau}$ composes an affine transformation with the analytic sine, and the time dependence enters solely through the parameter $\mathbf{b}_m(\tau)$, the overall function $\Phi_i(\tau)$ remains infinitely differentiable with respect to $\tau$, inheriting the smoothness guarantees of SIREN.
This formulation preserves the architectural simplicity of the original SIREN while seamlessly embedding temporal $\tau$ information into every hidden layer.

\begin{figure}
    \centering
    \includegraphics[width=1.0\linewidth]{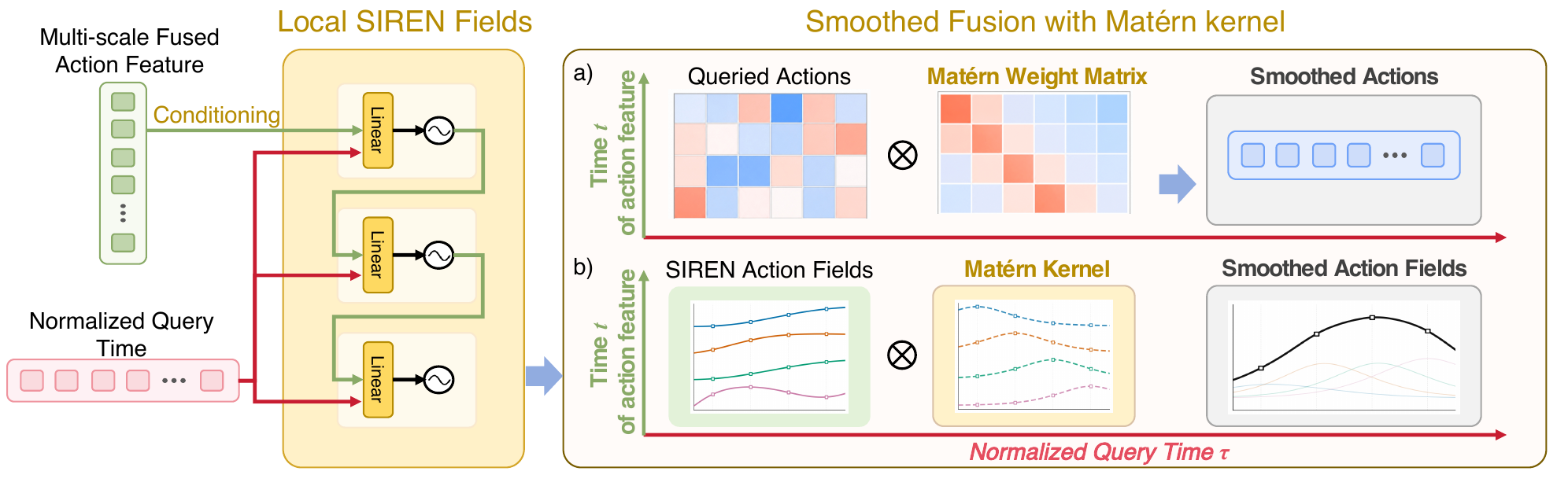}
    \caption{Overview of the conditional SIREN decoder. Local SIREN fields conditioned on multi-scale action features are queried over continuous time and fused with Matérn-kernel weights to generate temporally smooth actions.
}
    \vspace{-3mm}
    \label{fig:cond_siren}
\end{figure}

\textbf{Smooth Fusion with Mat\'ern Kernel.}
The global continuous function $\psi(\tau)$ is obtained by aggregating the local SIREN fields with a normalized Mat\'ern-$5/2$ kernel~\cite{williams2006gaussian} of length-scale parameter $\rho$ as
\begin{equation}
    \begin{gathered}
        \psi(\tau) = \sum_{i=1}^{H} \alpha_i(\tau)\,\Phi_i(\tau), \quad 
        \text{s.t.} \quad \alpha_i(\tau) \propto \kappa_{5/2}(|\tau-\tau_i|/\rho) .
    \end{gathered}
\end{equation}
This soft assignment encourages
second-order temporal smoothness across neighboring feature positions, reducing high-frequency inconsistencies while preserving local expressiveness.
The aggregated output $\{\psi(\tau_j)\}_{j=1}^{H}$ is interpreted as the predicted action chunk (regression) or as the estimated velocity field (flow matching), depending on the training objective of the base policy. 

\subsection{Training Objective}

ActionUNet is compatible with both direct regression and flow-matching policies. {We keep the pre-trained backbone frozen and only fine-tune the action expert with ActionUNet}, so the method can be trained under the original objective of the base policy.

\textbf{Regression Objective.}
For direct regression, we supervise the output sequence $\{\psi(\tau_j)\}_{j=1}^{H}$ with the ground-truth action chunk $\mathbf{a}_{1:H}$ using an $\ell_1$ loss:
\begin{equation}
\mathcal{L}_{\mathrm{reg}} = \frac{1}{H}\sum_{j=1}^{H}\big\|\psi(\tau_j) - \mathbf{a}_j\big\|_1.
\end{equation}

\textbf{Flow-matching Objective.}
For flow-based training, the decoder $\psi(\tau)$ is trained to predict the conditional vector field that transports 
noise to data.
Following the standard conditional flow-matching formulation~\cite{lipmanflow}, we construct a linear interpolation path between a noise sequence $\mathbf{\epsilon}_{1:H}$ and the clean action chunk $\mathbf{a}_{1:H}$.
Given an observation and the noisy trajectory $\mathbf{\epsilon}_{1:H}$, the model predicts $\psi(\tau_j)$, and we minimize the mean squared error against this velocity:
\begin{equation}
\mathcal{L}_{\mathrm{fm}} = \frac{1}{H}\sum_{j=1}^{H}\big\|\psi(\tau_j) - (\mathbf{a}_j - \mathbf{\epsilon}_j) \big\|_2^2.
\end{equation}

\textbf{Temporally Correlated Flow Noise.}
To align the flow target with the smoothness prior of our continuous decoder, we replace the standard white Gaussian noise with Matérn-$5/2$ process noise. Concretely, we sample noise vectors with a temporal covariance $K_{ij} = \kappa_{5/2}(|i-j|/\sigma) + \varepsilon\delta_{ij}$, where $\sigma$ controls the correlation length and $\varepsilon$ ensures numerical stability. This correlated perturbation reduces high-frequency jitter in the transport target and better matches the $C^2$ continuity induced by the SIREN decoder and Matérn aggregation.


\newcommand{\TBD}{\textcolor{red}{--}}
\newcommand{\pio}{$\pi_{0.5}$}

\section{Experiments}

\subsection{Experimental Setup}

\textbf{Simulation Benchmarks}. We evaluate \modelname{} on three simulation benchmarks:

1) \textbf{RoboTwin 2.0}~\cite{Robotwin} contains 50 bimanual tasks, each with easy and hard settings (varying distractors, backgrounds, lighting, table heights, and language). We select 12 tasks covering diverse temporal scale compositions. Each single-task policy is trained on 50 clean demonstrations and evaluated over 100 rollouts in both settings. 2) \textbf{LIBERO}~\cite{Libero} comprises four suites (Spatial, Object, Goal, Long), each with 10 tasks and 50 demonstrations. We train on a mixture from all suites. 3) \textbf{LIBERO-Plus}~\cite{Libero-plus} extends LIBERO with seven perturbation types (camera, robot state, language, lighting, background, sensor noise, layout). Models trained on LIBERO are directly evaluated on LIBERO-Plus to measure robustness. All tasks adopt the average success rate as the evaluation metric.

\textbf{Real-World Robot Experiments.}
We conduct real-world experiments on the Realman Gen72 dual-arm platform. Three Intel RealSense D435 cameras are used: two wrist-mounted cameras placed on the end-effectors and one head camera positioned between the two arms, providing multi-view observations of the workspace. We train and evaluate models on three dual-arm manipulation tasks: Place Object Basket, Stack Blocks Two, and Stack Bowls Two. To test real-world robustness, we consider two evaluation settings: an easy setting matching the training environment and a hard setting with additional distractor objects.




\textbf{Implementation Details.}
We apply \modelname{} to both \pio{}, a flow-based policy, and OpenVLA-OFT, a regression-based policy, to evaluate its compatibility with different VLA frameworks.
For \modelname{}, we set the temporal pyramid downsampling rate $r=2$, the Mat\'ern kernel length-scale $\rho=0.2$, and use standard SIREN initialization~\cite{SIREN}. We reproduce \pio{} following the official PyTorch version and OpenVLA-OFT baselines. All other baseline results are cited from~\cite{Robotwin, VLA-Adapter, Libero-plus}. All methods adopt the same dataset-specific training data, inference setting, and evaluation protocol. Unless otherwise specified, fine-tuning hyperparameters follow the settings reported in the original papers or official benchmark reports.

For RoboTwin 2.0, \pio{} and \pio{}+\modelname{} are trained on 2$\times$RTX 4090 GPUs for 8K steps with a batch size of 64 and a $5\times10^{-5}$ learning rate (cosine decay). For LIBERO and LIBERO-Plus, they are trained on 8$\times$A800 GPUs for 40K steps with a batch size of 256. OpenVLA-OFT and OpenVLA-OFT+\modelname{} follow official settings, except OpenVLA-OFT+\modelname{} on LIBERO uses a reduced learning rate of $5\times10^{-5}$ for stable fine-tuning.

\subsection{Results on RoboTwin 2.0}

Table~\ref{tab:robotwin_scale_grouped_results} reports the results on RoboTwin 2.0, where tasks are grouped according to their scale composition. Based on $\pi_{0.5}$, \modelname{} consistently improves performance across both easy and hard settings. Averaged over all 12 tasks, \modelname{} significantly improves the success rate by 8.8\% in the easy setting, and 9.8\% in the hard setting. The improvement under the hard setting is especially important, since this setting introduces stronger environmental perturbations and requires the policy to maintain reliable action execution under distractors.

The gains are consistent across scale compositions. On single-scale tasks, \modelname{} improves the average success rate from 66.0\% to 72.7\% and from 34.0\% to 44.0\% in the easy and hard settings, respectively. On dual-scale tasks, \modelname{} improves the average success rate from 60.2\% to 69.5\% and from 29.7\% to 40.0\% under easy and hard settings, respectively. On three-scale tasks, where successful execution requires coordinating more diverse temporal action patterns, \modelname{} improves the average success rate from 52.0\% to 62.0\% and from 25.7\% to 34.3\% in the easy and hard settings, respectively. These results show that multi-scale temporal action modeling is beneficial not only for simple action patterns but also for tasks involving more complex scale composition.

\begin{table*}[!t]
  \centering
  \caption{Task-wise success rates on RoboTwin 2.0 grouped by scale composition.
  }
  \resizebox{\linewidth}{!}{%
    \begin{tabular}{l*{8}{c}|*{2}{c}|*{2}{c}}
      \toprule
      \textbf{Task}
      & \multicolumn{2}{c}{\textbf{RDT~\cite{RDT}}}
      & \multicolumn{2}{c}{\textbf{$\pi_0$}~\cite{pi0}}
      & \multicolumn{2}{c}{\textbf{ACT}~\cite{ACT}}
      & \multicolumn{2}{c}{\textbf{DP3}~\cite{DP3}}
      & \multicolumn{2}{c}{\textbf{$\pi_{0.5}$}~\cite{pi05}}
      & \multicolumn{2}{c}{\cellcolor{orange!20}
\makebox[0pt][c]{\pio{}+\textbf{\modelname{}}}
      } \\
      & \textbf{Easy} & \textbf{Hard}
      & \textbf{Easy} & \textbf{Hard}
      & \textbf{Easy} & \textbf{Hard}
      & \textbf{Easy} & \textbf{Hard}
      & \textbf{Easy} & \textbf{Hard}
      & \cellcolor{orange!20}\textbf{Easy} & \cellcolor{orange!20}\textbf{Hard} \\

      \midrule
      \multicolumn{13}{c}{\textbf{Single-scale Tasks}} \\
      \midrule

      Pick Dual Bottles
      & 42 & 13
      & 57 & 12
      & 31 & 0
      & 60 & 1
      & 65 & 34
      & \cellcolor{orange!10}\textbf{71} & \cellcolor{orange!10}\textbf{39} \\

      Handover Mic
      & 90 & 31
      & 98 & 13
      & 85 & 0
      & \textbf{100} & 3
      & \textbf{100} & 55
      & \cellcolor{orange!10}\textbf{100} & \cellcolor{orange!10}\textbf{76} \\

      Handover Block
      & 45 & 14
      & 45 & 8
      & 42 & 0
      & \textbf{70} & 0
      & 33 & 13
      & \cellcolor{orange!10}47 & \cellcolor{orange!10}\textbf{17} \\

      \midrule
      \multicolumn{13}{c}{\textbf{Dual-scale Tasks}} \\
      \midrule

      Beat Block Hammer
      & 77 & 37
      & 43 & 21
      & 56 & 3
      & 72 & 8
      & 77 & 28
      & \cellcolor{orange!10}\textbf{89} & \cellcolor{orange!10}\textbf{43} \\

      Move Can Pot
      & 25 & 12
      & 58 & 21
      & 22 & 4
      & 70 & 6
      & 66 & 52
      & \cellcolor{orange!10}\textbf{77} & \cellcolor{orange!10}\textbf{69} \\

      Place A2B Left
      & 3 & 1
      & 31 & 1
      & 1 & 0
      & 46 & 2
      & 46 & 7
      & \cellcolor{orange!10}\textbf{55} & \cellcolor{orange!10}\textbf{10} \\

      Place Object Stand
      & 15 & 5
      & 36 & 11
      & 1 & 0
      & 60 & 0
      & 51 & 29
      & \cellcolor{orange!10}\textbf{62} & \cellcolor{orange!10}\textbf{38} \\

      Place Phone Stand
      & 15 & 6
      & 35 & 7
      & 2 & 0
      & 44 & 2
      & 47 & 21
      & \cellcolor{orange!10}\textbf{61} & \cellcolor{orange!10}\textbf{27} \\

      Press Stapler
      & 41 & 24
      & 62 & 29
      & 31 & 6
      & 69 & 3
      & \textbf{74} & 41
      & \cellcolor{orange!10}73 & \cellcolor{orange!10}\textbf{53} \\

      \midrule
      \multicolumn{13}{c}{\textbf{Three-scale Tasks}} \\
      \midrule

      Stack Blocks Two
      & 21 & 2
      & 42 & 1
      & 25 & 0
      & 24 & 0
      & 68 & 24
      & \cellcolor{orange!10}\textbf{74} & \cellcolor{orange!10}\textbf{35} \\

      Blocks Ranking RGB
      & 3 & 0
      & 19 & 5
      & 1 & 0
      & 3 & 0
      & 36 & 16
      & \cellcolor{orange!10}\textbf{50} & \cellcolor{orange!10}\textbf{24} \\

      Put Bottles Dustbin
      & 21 & 4
      & 54 & 13
      & 27 & 1
      & 60 & 21
      & 52 & 37
      & \cellcolor{orange!10}\textbf{62} & \cellcolor{orange!10}\textbf{44} \\

      \midrule
      \textbf{Average}
      & 33.2 & 12.4
      & 48.3 & 11.8
      & 27.0 & 1.2
      & 56.5 & 3.8
      & 59.6 & 29.8
      & \cellcolor{orange!10}\textbf{68.4}
      & \cellcolor{orange!10}\textbf{39.6} \\
      \bottomrule
    \end{tabular}
  }
  \vspace{-1.5mm}
  \label{tab:robotwin_scale_grouped_results}
\end{table*}

\begin{table*}[!t]
  \centering
  \caption{LIBERO benchmark results. We report the average success rate (\%) across four LIBERO task suites.}
  \footnotesize
  \setlength{\tabcolsep}{10pt}
    \begin{tabular}{lccccc}
      \toprule
      \textbf{Method} & \textbf{SPATIAL} & \textbf{OBJECT} & \textbf{GOAL} & \textbf{LONG} & \textbf{Avg.} \\
      \midrule

      Diffusion Policy~\cite{DP} 
      & 78.3 & 92.5 & 68.3 & 50.5 & 72.4 \\
      WorldVLA~\cite{WorldVLA}         
      & 87.6 & 96.2 & 83.4 & 60.0 & 81.8 \\
      SmolVLA~\cite{SmolVLA} 
      & 93.0 & 94.0 & 91.0 & 77.0 & 88.8 \\
      $\pi_{0}$~\cite{pi0}        
      & 96.8 & 98.8 & 95.8 & 85.2 & 94.2 \\
      $\pi_{0}$-FAST~\cite{pi0fast}   
      & 96.4 & 96.8 & 88.6 & 60.2 & 85.5 \\
      UniVLA~\cite{UniVLA}          
      & 96.5 & 96.8 & 95.6 & 92.0 & 95.2 \\
      VLA-Adapter~\cite{VLA-Adapter}      
      & 97.8 & 99.2 & 97.2 & 95.0 & 97.3 \\
      \midrule

      OpenVLA-OFT~\cite{OpenVLA-OFT}      
      & 98.4 & 99.0 & 98.4 & 95.8 & 97.9 \\
      \rowcolor{orange!20}
      OpenVLA-OFT+\textbf{ActionUNet} & \textbf{99.2} & 99.0 & \textbf{99.6} & \textbf{96.4} & \textbf{98.6} \\
\midrule

      $\pi_{0.5}$~\cite{pi05}      
      & 95.4 & 98.4 & 97.0 & 91.6 & 95.6 \\
      \rowcolor{orange!20}
      \textbf{$\pi_{0.5}$+ActionUNet} & 98.6 & \textbf{99.4} & 98.8 & 93.8 & 97.7 \\
      \bottomrule
    \end{tabular}%
  \vspace{-1.5mm}
  \label{tab:libero}
\end{table*}

\begin{table*}[!t]
  \centering
  \caption{Performance under environmental perturbations on LIBERO-Plus. All models are trained on LIBERO and evaluated on LIBERO-Plus. We report success rates (\%) under 7 perturbation types.}
  \footnotesize
    \begin{tabular}{lcccccccc}
      \toprule
      \textbf{Method} & \textbf{Camera} & \textbf{Robot} & \textbf{Lang.} & \textbf{Light} & \textbf{Back.} & \textbf{Noise} & \textbf{Layout} & \textbf{Avg.} \\
      \midrule

      OpenVLA~\cite{OpenVLA}
      & 0.8 & 3.5 & 23.0 & 8.1 & 34.8 & 15.2 & 28.5 & 15.6 \\

      WorldVLA~\cite{WorldVLA}
      & 0.1 & 27.9 & 41.6 & 43.7 & 17.1 & 10.9 & 38.0 & 25.0 \\

      UniVLA~\cite{UniVLA}
      & 1.8 & 46.2 & 69.6 & 69.0 & 81.0 & 21.2 & 31.9 & 43.9 \\

      $\pi_{0}$~\cite{pi0}
      & 13.8 & 6.0 & 58.8 & 85.0 & 81.4 & \textbf{79.0} & 68.9 & 53.6 \\

      $\pi_{0}$-FAST~\cite{pi0fast}
      & \textbf{65.1} & 21.6 & 61.0 & 73.2 & 73.2 & 74.4 & 68.8 & 61.6 \\

      \midrule

      OpenVLA-OFT~\cite{OpenVLA-OFT}
      & 36.0 & 35.9 & 67.0 & 82.1 & 93.1 & 47.1 & 82.5 & 60.9 
      \\

      \rowcolor{orange!20}
      OpenVLA-OFT+\textbf{ActionUNet}
      & 41.2 & 40.8 & 68.3 & 83.4 & \textbf{93.7} & 54.5 & 85.3 & 64.6 \\
      \midrule

      $\pi_{0.5}$~\cite{pi05}
      & 48.2 & 45.9 & 68.8 & 93.0 & 87.4 & 52.0 & 81.7 & 65.8 \\

      \rowcolor{orange!20}
      \textbf{$\pi_{0.5}$+ActionUNet}
      & 56.9 & \textbf{54.4} & \textbf{73.3} & \textbf{94.6} & 89.8 & 61.1 & \textbf{85.7} & \textbf{71.9} \\
      \bottomrule
    \end{tabular}%
  \vspace{-2.5mm}
  \label{tab:perturbation_results}
\end{table*}

\subsection{Results on LIBERO and LIBERO-Plus}

Table~\ref{tab:libero} reports results on the standard LIBERO benchmark.
With $\pi_{0.5}$, \modelname{} improves the performance of all four suites, achieving an average success rate of 97.7\% against 95.6\%.
This improvement extends to the regression-based OpenVLA-OFT backbone (98.6\% vs 97.9\%), confirming that \modelname{} generalizes across VLA models while preserving strong in-distribution performance.




Table~\ref{tab:perturbation_results} evaluates the robustness of \modelname{} and comparison methods under controlled perturbations on LIBERO-Plus.
$\pi_{0.5}$+\modelname{} improves the average success rate from 65.8\% to 71.9\%, with consistent gains across all seven perturbation types, particularly for camera, robot-state, and sensor-noise perturbations.
This result demonstrates that the temporal U-Net refines action features for multi-scale representations, and the continuous decoder reduces high-frequency inconsistency to benefit the robotic execution.
With the more powerful regression-based OpenVLA-OFT, \modelname{} still improves the average success rate from 
60.9\% to 64.6\%,
indicating its robustness across VLA frameworks.


\subsection{Real-World Experiments}


\textbf{Evaluation Protocol.}
We select three representative tasks from the RoboTwin 2.0 benchmark: place Object Basket, Stack Blocks Two, and Stack Bowls Two.
Each task contains easy and hard evaluation settings. 
The easy setting is the same as the training settings. 
For the hard setting, we randomly add some distractor objects to the environment to introduce visual and physical interference.
We report the success rate as the percentage of successful trials out of the 50 rollouts.

\textbf{Results.}
Table~\ref{tab:baseline_vs_ours} reports the real-world evaluation results on the Realman {Robot}. \modelname{} consistently improves the baseline across all three tasks under both easy and hard evaluation settings. The average success rate is significantly improved from 76.0\% to 88.7\% in the easy setting and from 57.3\% to 68.7\% in the hard setting.
These results demonstrate its robustness in complex real-world environments without compromising performance in standard scenes.

\begin{table}[!t] 
  \centering
  \caption{Comparison of our method against the baseline under easy and hard conditions in real-world experiments.}
  \vspace{1mm}
    \footnotesize
    \begin{tabular}{l c c c c}
      \toprule
      \multirow{2}{*}{\textbf{Task}} 
      & \multicolumn{2}{c}{\textbf{\pio}}
      & \multicolumn{2}{c}{\makebox[0pt][c]{\pio+\textbf{\modelname{}}}} \\
      \cmidrule(lr){2-3} \cmidrule(lr){4-5}
      & \textbf{Easy} & \textbf{Hard} & \textbf{Easy} & \textbf{Hard} \\
      \midrule
      Place Object Basket & 60 & 42 & 78 & 52 \\
      Stack Bowls Two     & 88 & 62 & 96 & 74 \\
      Stack Blocks Two    & 80 & 68 & 92 & 80 \\
      \midrule
      Average    & 76.0 & 57.3 & \textbf{88.7} & \textbf{68.7} \\
      \bottomrule
    \end{tabular}%
    \vspace{-1.5mm}
  \label{tab:baseline_vs_ours}
\end{table}

\definecolor{easybg}{RGB}{238,247,255}
\definecolor{hardbg}{RGB}{255,244,232}


\newcolumntype{L}[1]{>{\raggedright\arraybackslash}p{#1}}
\newcolumntype{C}[1]{>{\centering\arraybackslash}p{#1}}

\begin{table*}[!t]
  \centering
  \caption{Ablation results on four RoboTwin 2.0 tasks. 
  Each task reports success rates under easy and hard settings. All non-baseline variants are built on $\pi_{0.5}$.}
  \vspace{-1.5mm}
  {\footnotesize
  \renewcommand{\arraystretch}{1.15}
  \setlength{\tabcolsep}{2pt}
    \begin{tabular}{@{}L{0.19\textwidth}*{10}{C{0.060\textwidth}}@{}}
      \toprule
      \multirow{2}{*}{\textbf{Task}} 
      & \multicolumn{2}{c}{\textbf{$\pi_{0.5}$}}
      & \multicolumn{2}{c}{\textbf{+ U-Net}}
      & \multicolumn{2}{c}{\textbf{+ SIREN}}
      & \multicolumn{2}{c}{\makecell[c]{\textbf{+ \modelname{}}\\\textbf{w/o Matérn}}}
      & \multicolumn{2}{c}{\textbf{+ \modelname{}}} \\
      \cmidrule(lr){2-3}
      \cmidrule(lr){4-5}
      \cmidrule(lr){6-7}
      \cmidrule(lr){8-9}
      \cmidrule(lr){10-11}
      & \textbf{Easy} & \textbf{Hard} 
      & \textbf{Easy} & \textbf{Hard} 
      & \textbf{Easy} & \textbf{Hard} 
      & \textbf{Easy} & \textbf{Hard}
      & \textbf{Easy} & \textbf{Hard} \\
      \midrule

      Move Can Pot 
      & 66 & 52 
      & 64 & 56 
      & 73 & 53 
      & 73 & 68
      & \textbf{77} & \textbf{69} \\

      Stack Blocks Two 
      & 68 & 24 
      & \textbf{75} & 27 
      & 74 & 24 
      & 71 & 34
      & 74 & \textbf{35} \\

      Beat Block Hammer 
      & 77 & 28 
      & 65 & 33 
      & 87 & 30 
      & \textbf{89} & 39
      & \textbf{89} & \textbf{43} \\

      Pick Dual Bottles 
      & 65 & 34 
      & 66 & 33 
      & 68 & 26 
      & 67 & \textbf{40}
      & \textbf{71} & 39 \\

      \midrule
      \rowcolor{gray!8}
      Average
      & 69.0 & 34.5
      & 67.5 & 37.3
      & 75.5 & 33.3
      & 75.0 & 45.3
      & \textbf{77.8} & \textbf{46.5} \\

      Gain over $\pi_{0.5}$
      & -- & --
      & -1.5 & +2.8
      & +6.5 & -1.2
      & +6.0 & +10.8
      & \textbf{+8.8} & \textbf{+12.0} \\
      \bottomrule
    \end{tabular}%
  }
  \vspace{-1.5mm}
  \label{tab:ablation_results}
\end{table*}

\subsection{Ablation Study}

\begin{figure}[htbp]
    \centering
    \includegraphics[width=0.99\linewidth]{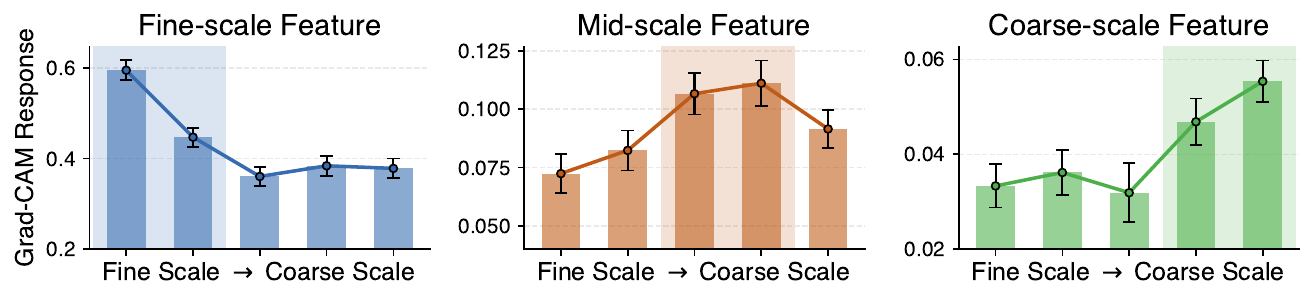}
    \caption{
    Scale-conditioned temporal Grad-CAM responses on RoboTwin 2.0.
    Fine-, mid-, and coarse-resolution temporal features show stronger responses to the corresponding action scales.
    }
    \vspace{-3mm}
    \label{fig:gradcam}
\end{figure}

\textbf{Module Ablation.}
We ablate \modelname ~with respect to the temporal U-Net, the conditional SIREN decoder, and the Matérn-weighted aggregation in Table~\ref{tab:ablation_results}.
Using only the temporal U-Net on $\pi_{0.5}$ improves hard-setting performance from 34.5\% to 37.3\% while slightly dropping in the easy setting. 
This suggests that while the temporal U-Net enhances robustness against distractors, it compromises the spatial continuity of actions.
In contrast, by adding only the SIREN decoder, the success rate in the easy setting is improved from 69.0\% to 75.5\% while dropping in the hard setting. It indicates the SIREN decoder is capable of maintaining the spatial continuity of actions to suppress trajectory jitter, but fails to capture multi-scale action representations, leading to a lack of robustness against distractors.
The combination of U-Net and SIREN without the Matérn kernel (ActionUNet w/o Matérn) achieves 75.0\% and 45.3\% success rates in easy and hard settings, significantly outperforming the baseline by 6.0\% and 10.8\%, demonstrating that multi-scale modeling, aided by SIREN to mitigate action discontinuity, effectively provides crucial robustness against distractors.
Incorporating the Matérn kernel further increases performance to 77.8\% and 46.5\% in easy and hard settings, suggesting that the Matérn kernel supplies a lightweight smoothness prior over neighboring local fields, complementing the learned components.
\textbf{Effect of Multi-scale Fine-tuning.} We compute the response of the 1st, 4th, and 6th layers of temporal U-Net to reflect fine-, mid-, and coarse-scale action feature activations using Grad-CAM on RoboTwin 2.0. The responses are grouped into five bins by ground-truth action scales, and Grad-CAM responses are min-max normalized and averaged per bin.
As shown in Fig.~\ref{fig:gradcam}, action features at each scale respond most strongly to the corresponding ground-truth action scales, suggesting that ActionUNet learns scale-aware temporal representations.

\begin{figure}[htbp]
    \centering
    \begin{minipage}[t]{0.48\textwidth}
        \centering
        \vspace{0pt}
        \includegraphics[width=\linewidth]{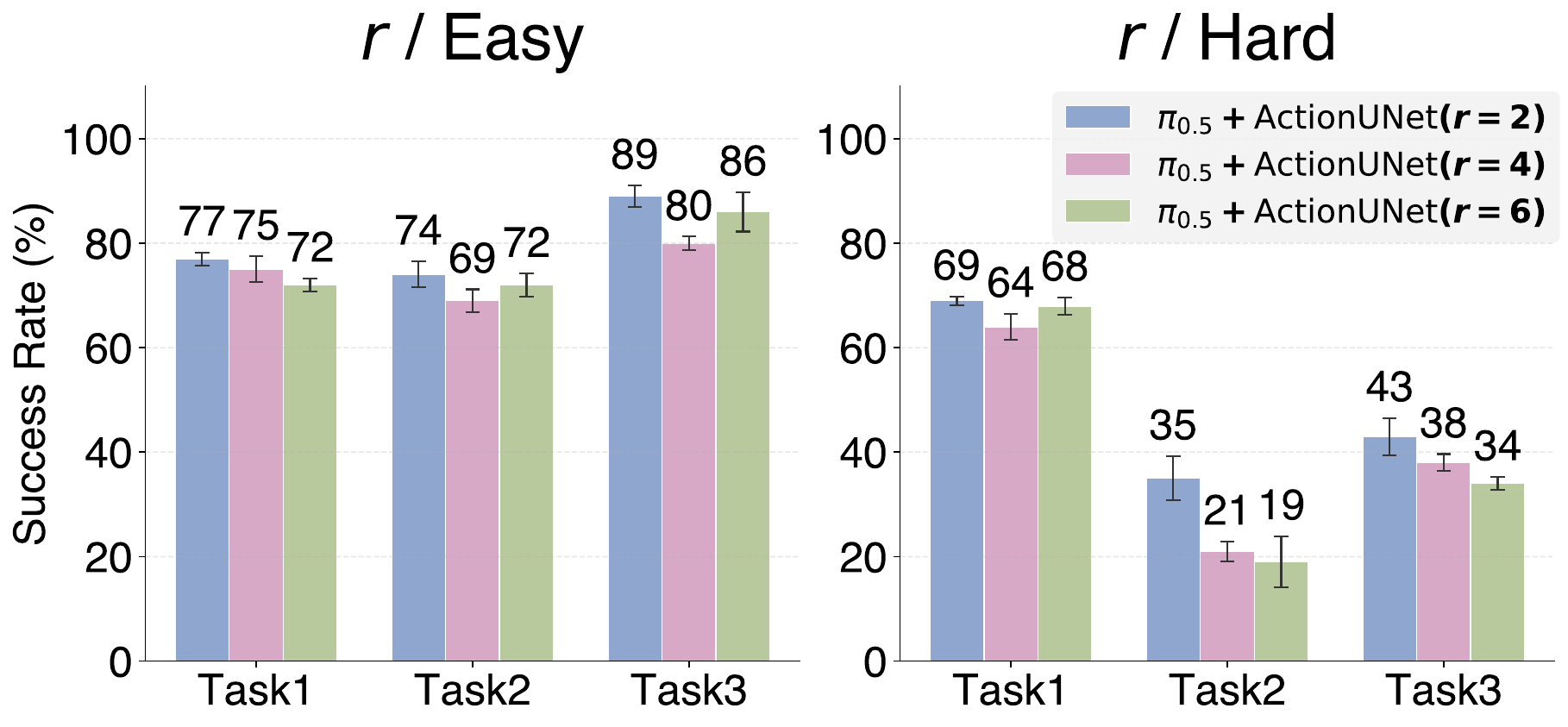}
        \caption{
        Ablation study on the downsampling rate $r$. Task1, Task2, and Task3 correspond to Move Can Pot, Stack Blocks Two, and Beat Block Hammer, respectively.
        }
        \vspace{-3mm}
        \label{fig:ablation}
    \end{minipage}
    \hfill
    \begin{minipage}[t]{0.48\textwidth}
        \centering
        \vspace{0pt}
        \includegraphics[width=\linewidth]{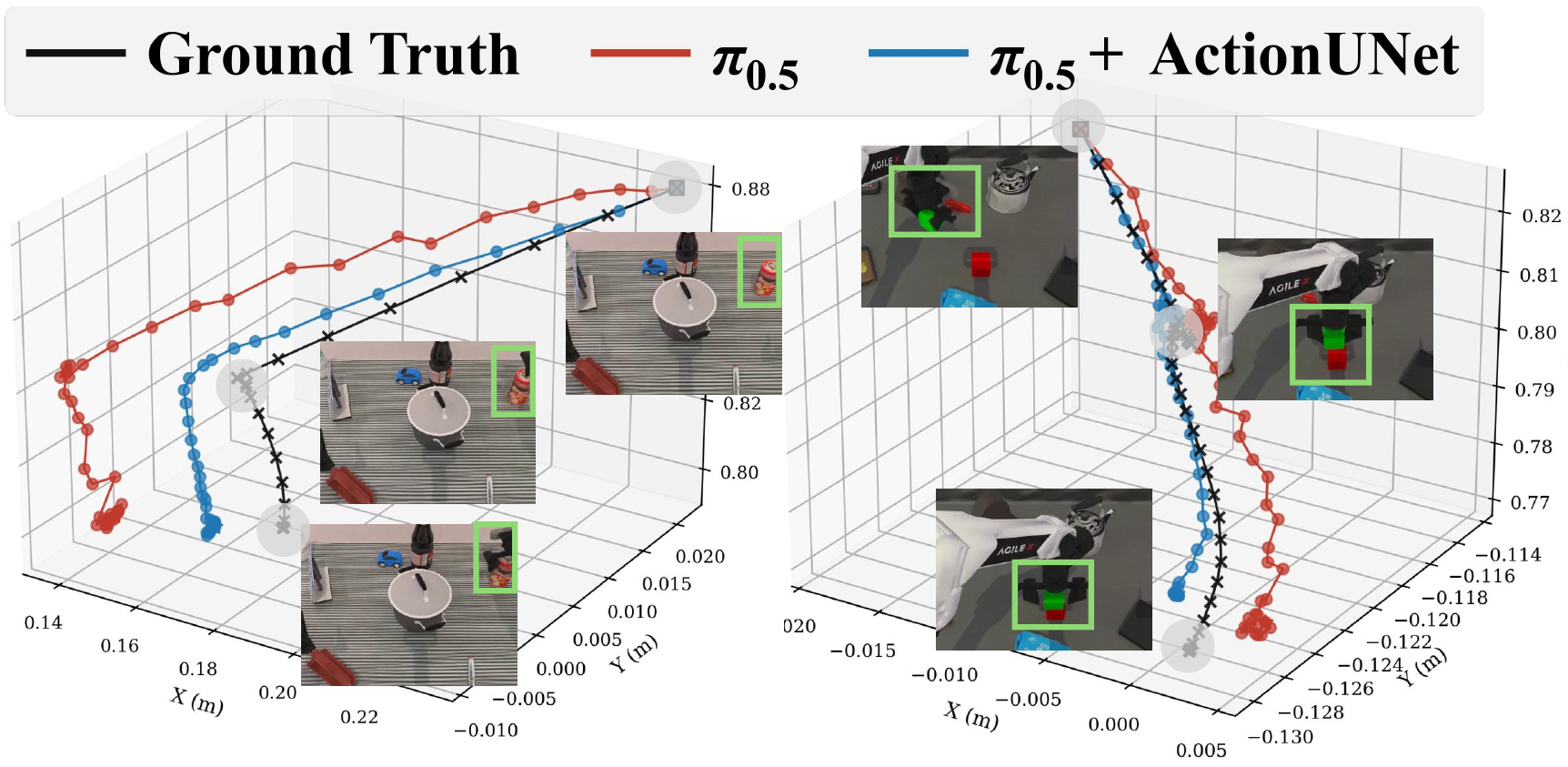}
        \caption{
        Predicted end-effector action chunks under the same initial settings.
        ActionUNet produces actions closer to the ground-truth with smoother local action than $\pi_{0.5}$.
        }
        \vspace{-3mm}
        \label{fig:vis_result}
    \end{minipage}
\end{figure}

\textbf{Effect of Hyper-parameters $r$.}
Fig.~\ref{fig:ablation} shows that the downsampling rate $r=2$ in the temporal U-Net performs best under both easy and hard settings.
Smaller $r$ provides denser intermediate scales for gradual coarse-to-fine refinement, which is especially beneficial in hard settings where contact-sensitive actions are required.

\textbf{Visualization Analysis.}
As shown in Fig.~\ref{fig:vis_result}, we further visualize predicted end-effector action chunks under the same initial setting.
Compared with $\pi_{0.5}$, ActionUNet produces smaller spatial deviation and smoother trajectories in these examples. This demonstrates ActionUNet's effectiveness in generating precise and stable continuous actions.

\section{Conclusion}

We propose ActionUNet, a lightweight multi-scale fine-tuning module for temporal action mapping in pre-trained VLA policies. 
ActionUNet injects hierarchical temporal priors into the action feature space through a temporal U-Net, and decodes the refined features with a conditional SIREN decoder to preserve smooth action trajectories. This design provides an architecture-compatible interface for both flow-based and regression-based policies, while introducing negligible computational overhead. On RoboTwin 2.0, LIBERO, LIBERO-Plus, and real-world dual-arm experiments, ActionUNet consistently improves task success and robustness under environmental distractions, showing the effectiveness of multi-scale temporal mapping for efficient VLA adaptation.

\textbf{Limitations and Future Work.}

ActionUNet currently uses fixed temporal scales, models action chunks
independently, and introduces scale-aware modeling only at downstream
action stage. Future work will explore adaptive scales, cross-chunk temporal
modeling, and scale-aware VLA pre-training.

\begin{ack}
This work was supported by the National Natural Science Foundation of China (NSFC) under Grant 62472402 and 62176260, and the Fundamental Research Funds for the Central Universities (E2ET1104).
The authors declare no competing interests.
\end{ack}

\newpage

\bibliographystyle{IEEEtran} 
\bibliography{main}

\newpage

\appendix

\section{Additional Method Details}
\label{app:method_details}

\textbf{Conditional SIREN Decoder.}
Given the refined action features $\tilde{\mathbf{h}}^{(0)}_1,\dots,\tilde{\mathbf{h}}^{(0)}_H$, ActionUNet represents an action chunk as samples from a continuous function over the normalized time axis $\tau\in[-1,1]$. The temporal anchors are uniformly placed as
\begin{equation}
\tau_i=-1+\frac{2(i-1)}{H-1},
\qquad i=1,\ldots,H.
\end{equation}
For each anchor $i$, the local SIREN field $\Phi_i(\tau)$ maps a query time $\tau$ to the action space while conditioning on the refined action feature $\tilde{\mathbf{h}}^{(0)}_i$. The final continuous function is obtained by Mat\'ern-weighted aggregation:
\begin{equation}
\psi(\tau)=\sum_{i=1}^{H}\alpha_i(\tau)\Phi_i(\tau),
\qquad
\alpha_i(\tau)=
\frac{\kappa_{5/2}(|\tau-\tau_i|/\rho)}
{\sum_{j=1}^{H}\kappa_{5/2}(|\tau-\tau_j|/\rho)}.
\end{equation}
The output sequence $\{\psi(\tau_j)\}_{j=1}^{H}$ is interpreted as the predicted action chunk in regression mode, and as the estimated flow-matching velocity field in flow mode.

\textbf{Smoothness of Mat\'ern-weighted Decoding.}
We verify that Mat\'ern-$5/2$ weighting preserves second-order smoothness of the decoded function. Let $c=\sqrt{5}/\rho$. The kernel and its derivative are
\begin{equation}
\begin{aligned}
g_\rho(t)
&=\left(1+ct+\frac{c^2t^2}{3}\right)e^{-ct},\\
g_\rho'(t)
&=-\frac{c^2}{3}t(1+ct)e^{-ct}.
\end{aligned}
\end{equation}
For each anchor, define $q_i(\tau)=g_\rho(|\tau-\tau_i|)$. Since $g_\rho'(0)=0$ and $g_\rho''(0)=-c^2/3$ is finite, the left and right first and second derivatives of $q_i$ agree at $\tau_i$. Hence $q_i(\tau)$ is $C^2$ on $[-1,1]$.

Since $q_j(\tau)>0$ for all $j$ and $\tau$, the denominator $Z(\tau)=\sum_{j=1}^{H}q_j(\tau)$ is strictly positive. Therefore, each normalized weight $\alpha_i(\tau)=q_i(\tau)/Z(\tau)$ is also $C^2$. Each local SIREN field $\Phi_i(\tau)$ is smooth in $\tau$, because it is composed of affine maps and sinusoidal nonlinearities. Consequently,
\begin{equation}
\psi(\tau)\in C^2([-1,1];\mathbb{R}^{D}).
\end{equation}

For implementation on the uniform temporal grid, the normalized-time length-scale $\rho$ is equivalent to an index-space length-scale $\rho_{\mathrm{idx}}=\rho/\Delta_\tau$, where $\Delta_\tau=2/(H-1)$.

\textbf{Temporally Correlated Noise for Flow Matching.}
For flow-based training, we use temporally correlated Gaussian noise on the action grid. The temporal covariance matrix is defined as
\begin{equation}
K_{ij}
=
\kappa_{5/2}\!\left(\frac{|i-j|}{\sigma}\right)
+\varepsilon\delta_{ij},
\end{equation}
where $\sigma$ controls the correlation length and $\varepsilon$ ensures numerical stability. Given the Cholesky decomposition $K=\mathbf{R}\mathbf{R}^{\top}$, we sample
\begin{equation}
\boldsymbol{\epsilon}_{1:H}
=
\mathbf{R}\boldsymbol{\eta}_{1:H},
\qquad
\boldsymbol{\eta}_{1:H}\sim\mathcal{N}(0,I).
\end{equation}
For multi-dimensional actions, this temporal covariance is applied independently to each action dimension, equivalently sampling $\boldsymbol{\epsilon}_{1:H}\sim\mathcal{N}(0,K\otimes I_D)$.

We denote the flow interpolation time by $s\in[0,1]$ to avoid overloading the normalized action time $\tau$. The linear interpolation path and target velocity are
\begin{equation}
\begin{aligned}
\mathbf{x}_s
&=(1-s)\boldsymbol{\epsilon}_{1:H}
+s\mathbf{a}_{1:H},\\
\mathbf{u}_s
&=\frac{d\mathbf{x}_s}{ds}
=\mathbf{a}_{1:H}-\boldsymbol{\epsilon}_{1:H}.
\end{aligned}
\end{equation}
The flow-mode decoder predicts this target velocity at each temporal query $\tau_j$. This construction aligns the flow target with the second-order temporal prior used by the continuous decoder, instead of injecting temporally white perturbations into the action sequence.

\textbf{Action Representation and Normalization.}
\modelname{} inherits the action representation of the backbone VLA model without modification. For $\pi_{0.5}$, actions are represented as continuous end-effector poses in the flow-matching framework. For OpenVLA-OFT, actions are represented as continuous joint/task-space commands in a regression framework.

During preprocessing, actions are normalized to zero mean and unit variance. The discrete action horizon $H$ is mapped to the continuous normalized time axis $\tau\in[-1,1]$ for SIREN decoding, improving training stability and gradient flow through the sinusoidal layers.

\section{Inference Efficiency}
\label{app:inference_efficiency}

ActionUNet is designed as a lightweight module for pre-trained VLA policies. We report policy-level inference latency to quantify the computational overhead introduced by the proposed module.

\textbf{Measurement Protocol.}
We measure the latency of sampling one action chunk on a single NVIDIA RTX 4090 GPU. Both $\pi_{0.5}$ and $\pi_{0.5}$+\textbf{ActionUNet} use the same input observation, action horizon of 50, and 10 flow-matching steps. All measurements are conducted under the same PyTorch inference setting with \texttt{torch.compile} enabled. We report the mean policy inference latency with 95\% confidence intervals computed over repeated policy-call measurements.

These measurements correspond to policy computation only. They should not be interpreted as end-to-end robot execution latency, which is also affected by mechanical motion, low-level controller constraints, safety-limited velocity, gripper actuation, and contact-rich interactions.

\begin{table}[!t]
\centering
\caption{
Compiled policy-call latency on a single RTX 4090 GPU. We report mean latency with 95\% confidence intervals computed over repeated policy-call measurements.
}
\vspace{3mm}
\label{tab:app_policy_latency}
\footnotesize
\renewcommand{\arraystretch}{1.12}
\setlength{\tabcolsep}{5pt}
\begin{tabular}{lccc}
\toprule
Metric & $\pi_{0.5}$ & $\pi_{0.5}$+\textbf{ActionUNet} & Difference \\
\midrule
Mean latency / policy call (ms)
& $60.947 \pm 0.251$
& $64.886 \pm 0.376$
& $+3.939 \quad (+6.46\%)$ \\
\bottomrule
\end{tabular}
\end{table}

\textbf{Rollout-level Policy Computation.}
We compute the average number of taken action steps over successful rollouts on the same 12 RoboTwin2.0 main experimental tasks used in the task-wise success-rate evaluation in Table~1. We then estimate cumulative policy inference latency using the compiled policy-call latency in Table~\ref{tab:app_policy_latency}. This metric measures policy computation only and excludes physical robot execution time.

\begin{table}[!t]
\centering
\caption{
Rollout-level policy computation on the 12 RoboTwin2.0 tasks. Average taken steps are task-level unweighted means over successful rollouts. Latency values are estimated by multiplying the average taken steps by the compiled policy-call latency in Table~\ref{tab:app_policy_latency}; their 95\% confidence intervals are propagated from the task-independent policy-call latency measurements. The paired difference is computed as $\pi_{0.5}$+\textbf{ActionUNet} minus $\pi_{0.5}$ latency, and its 95\% confidence interval is propagated from the two latency confidence intervals.
}
\label{tab:app_success_rollout_inference}
\footnotesize
\renewcommand{\arraystretch}{1.12}
\setlength{\tabcolsep}{2.5pt}
\begin{tabular}{lccccc}
\toprule
\multirow{2}{*}{Setting}
& \multicolumn{2}{c}{Average Taken Steps}
& \multicolumn{3}{c}{Latency (s)} \\
\cmidrule(lr){2-3}
\cmidrule(lr){4-6}
& $\pi_{0.5}$
& $\pi_{0.5}$+\textbf{ActionUNet}
& $\pi_{0.5}$
& $\pi_{0.5}$+\textbf{ActionUNet}
& Difference \\
\midrule
Easy
& 238.80
& 238.69
& $14.554 \pm 0.060$
& $15.488 \pm 0.090$
& $+0.934 \pm 0.108$ \\
Hard
& 276.43
& 264.29
& $16.848 \pm 0.069$
& $17.149 \pm 0.099$
& $+0.301 \pm 0.121$ \\
\bottomrule
\end{tabular}
\end{table}

As shown in Table~\ref{tab:app_success_rollout_inference}, $\pi_{0.5}$+\textbf{ActionUNet} introduces only a small rollout-level increase in policy computation. In the Easy setting, the cumulative policy-computation latency increases by 0.934 seconds. In the Hard setting, $\pi_{0.5}$+\textbf{ActionUNet} uses fewer average taken steps than the baseline, which partially offsets its higher per-call latency; the cumulative policy-computation latency increases by only 0.301 seconds. These results indicate that the additional policy-computation overhead remains small relative to the full successful rollout.


\section{Action Scale Computation}
\label{sec:scale_analysis_50}

We quantify task-level and stage-level action scales using per-step end-effector displacement to characterize temporal dynamics and group RoboTwin tasks.

\subsection{End-effector Displacement}

Let $\mathbf{p}^{L}_{t},\mathbf{p}^{R}_{t}\in\mathbb{R}^{3}$ denote the Cartesian positions of the left and right end-effectors at recorded step $t$. The per-step displacement is defined as
\[
d_t
=
\left\|
\mathbf{p}^{L}_{t+1} - \mathbf{p}^{L}_{t}
\right\|_2
+
\left\|
\mathbf{p}^{R}_{t+1} - \mathbf{p}^{R}_{t}
\right\|_2 .
\]
Displacement is measured in m/step rather than m/s, since demonstrations are generated from planned waypoints under a fixed recording protocol. It serves as a dataset-level action-scale proxy rather than physical execution velocity.

The overall task-level scale is averaged over valid consecutive steps from successful episodes:
\[
D_{\mathrm{task}}(\mathcal{T})
=
\frac{1}{|\mathcal{I}_{\mathcal{T}}|}
\sum_{t \in \mathcal{I}_{\mathcal{T}}}
d_t ,
\]
where $\mathcal{I}_{\mathcal{T}}$ denotes the set of valid step transitions from task $\mathcal{T}$.

\subsection{Stage-level Scale Regimes}

We segment trajectories into stages according to stable binary gripper-state transitions ($0$ and $1$), ignoring intermediate transition states. The stage-level scale is
\[
D_{\mathrm{stage}}(s)
=
\frac{1}{|\mathcal{I}_{s}|}
\sum_{t \in \mathcal{I}_{s}}
d_t ,
\]
where $\mathcal{I}_{s}$ contains valid step transitions within stage $s$.

\subsection{Scale Thresholds}

We use empirical tertile thresholds to classify task-level and non-idle stage-level displacements into low-, medium-, and high-scale regimes.

The task-level thresholds, computed over all 50 tasks, are
\[
\gamma^{\mathrm{task}}_1 = 0.03130,
\qquad
\gamma^{\mathrm{task}}_2 = 0.03714 .
\]
The task-level scale label is
\[
c_{\mathrm{task}}(\mathcal{T})
=
\begin{cases}
\text{Low}, & D_{\mathrm{task}}(\mathcal{T}) < \gamma^{\mathrm{task}}_1, \\
\text{Medium}, & \gamma^{\mathrm{task}}_1 \le D_{\mathrm{task}}(\mathcal{T}) < \gamma^{\mathrm{task}}_2, \\
\text{High}, & D_{\mathrm{task}}(\mathcal{T}) \ge \gamma^{\mathrm{task}}_2.
\end{cases}
\]

The stage-level thresholds, computed over all non-idle stages, are
\[
\gamma^{\mathrm{stage}}_1 = 0.01998,
\qquad
\gamma^{\mathrm{stage}}_2 = 0.04693 .
\]
The stage-level scale label is
\[
c_{\mathrm{stage}}(s)
=
\begin{cases}
\text{Low}, & D_{\mathrm{stage}}(s) < \gamma^{\mathrm{stage}}_1, \\
\text{Medium}, & \gamma^{\mathrm{stage}}_1 \le D_{\mathrm{stage}}(s) < \gamma^{\mathrm{stage}}_2, \\
\text{High}, & D_{\mathrm{stage}}(s) \ge \gamma^{\mathrm{stage}}_2.
\end{cases}
\]

The number of distinct active scale regimes in task $\mathcal{T}$ is
\[
K(\mathcal{T})
=
\left|
\left\{
c_{\mathrm{stage}}(s)
:
s \in \mathcal{T},
\;
D_{\mathrm{stage}}(s) > 0
\right\}
\right|.
\]
Tasks with $K(\mathcal{T})=1$, $2$, or $3$ are categorized as single-scale, dual-scale, or three-scale, respectively.

\subsection{Scale Analysis on 50 Tasks}

Table~\ref{tab:scale_analysis_50_tasks} reports the scale analysis of all 50 RoboTwin tasks.

\setlength{\LTcapwidth}{\textwidth}

\begingroup
\small
\setlength{\tabcolsep}{4pt}
\renewcommand{\arraystretch}{1}

\setlength{\LTcapwidth}{\textwidth}

\setlength{\LTleft}{0pt plus 1fill}
\setlength{\LTright}{0pt plus 1fill}

\begin{longtable}{p{3.2cm} c c p{3.1cm} c c}

\caption{Action scale analysis on 50 RoboTwin tasks. Overall displacement is measured in m/step. L, M, and H denote low-, medium-, and high-scale stages. Stage sequences exclude idle stages; composition lists distinct active regimes.}
\vspace{0mm}
\label{tab:scale_analysis_50_tasks}
\\
\toprule
\textbf{Task}
& \textbf{Overall}
& \textbf{Overall Scale}
& \textbf{Stage Sequence}
& \textbf{Composition}
& \textbf{\#Regimes} \\
\midrule
\endfirsthead

\toprule
\textbf{Task}
& \textbf{Overall}
& \textbf{Overall Scale}
& \textbf{Stage Sequence}
& \textbf{Composition}
& \textbf{\#Regimes} \\
\midrule
\endhead

\bottomrule
\endlastfoot

Adjust Bottle & 0.0442 & High & H--M & M+H & 2 \\
Beat Block Hammer & 0.0372 & High & H--L & L+H & 2 \\
Blocks Ranking RGB & 0.0363 & Medium & M--L--H--L--H--L--L & L+M+H & 3 \\
Blocks Ranking Size & 0.0359 & Medium & H--L--H--L--H--L--L & L+H & 2 \\
Click Alarmclock & 0.0412 & High & H--M & M+H & 2 \\
Click Bell & 0.0373 & High & H--L & L+H & 2 \\
Dump Bin Bigbin & 0.0335 & Medium & M--L--H--L & L+M+H & 3 \\
Grab Roller & 0.0675 & High & H--M & M+H & 2 \\
Handover Block & 0.0348 & Medium & M--M--M & M & 1 \\
Handover Mic & 0.0281 & Low & M--M--M & M & 1 \\
Hanging Mug & 0.0382 & High & H--L--H--L--L & L+H & 2 \\
Lift Pot & 0.0570 & High & H--M & M+H & 2 \\
Move Can Pot & 0.0348 & Medium & H--L & L+H & 2 \\
Move Pillbottle Pad & 0.0346 & Medium & H--M & M+H & 2 \\
Move Playingcard Away & 0.0361 & Medium & H--M & M+H & 2 \\
Move Stapler Pad & 0.0276 & Low & M--L & L+M & 2 \\
Open Laptop & 0.0179 & Low & M--L & L+M & 2 \\
Open Microwave & 0.0116 & Low & H--L--M--L & L+M+H & 3 \\
Pick Diverse Bottles & 0.0736 & High & H--H & H & 1 \\
Pick Dual Bottles & 0.0730 & High & H--H & H & 1 \\
Place A2B Left & 0.0307 & Low & M--L & L+M & 2 \\
Place A2B Right & 0.0299 & Low & M--M & M & 1 \\
Place Bread Basket & 0.0370 & Medium & H--M--M--L & L+M+H & 3 \\
Place Bread Skillet & 0.0438 & High & H--M & M+H & 2 \\
Place Burger Fries & 0.0440 & High & H--M--M--L & L+M+H & 3 \\
Place Can Basket & 0.0460 & High & H--M--H--L & L+M+H & 3 \\
Place Cans Plasticbox & 0.0554 & High & H--M--H--H & M+H & 2 \\
Place Container Plate & 0.0301 & Low & H--L--L & L+H & 2 \\
Place Dual Shoes & 0.0556 & High & H--M--H & M+H & 2 \\
Place Empty Cup & 0.0249 & Low & M--L--L & L+M & 2 \\
Place Fan & 0.0312 & Low & H--L & L+H & 2 \\
Place Mouse Pad & 0.0269 & Low & M--L & L+M & 2 \\
Place Object Basket & 0.0460 & High & H--L--H--L & L+H & 2 \\
Place Object Scale & 0.0316 & Medium & M--M & M & 1 \\
Place Object Stand & 0.0306 & Low & H--L & L+H & 2 \\
Place Phone Stand & 0.0279 & Low & M--L & L+M & 2 \\
Place Shoe & 0.0267 & Low & M--L & L+M & 2 \\
Press Stapler & 0.0296 & Low & H--L & L+H & 2 \\
Put Bottles Dustbin & 0.0384 & High & M--L--H--M--M--M--H--M--L--H--H & L+M+H & 3 \\
Put Object Cabinet & 0.0315 & Medium & M--M--M & M & 1 \\
Rotate QRcode & 0.0237 & Low & M--L & L+M & 2 \\
Scan Object & 0.0448 & High & H--M & M+H & 2 \\
Shake Bottle Horizontally & 0.0188 & Low & H--L & L+H & 2 \\
Shake Bottle & 0.0213 & Low & H--L & L+H & 2 \\
Stack Blocks Three & 0.0345 & Medium & M--L--H--L--H--L--L & L+M+H & 3 \\
Stack Blocks Two & 0.0333 & Medium & M--L--H--L--L & L+M+H & 3 \\
Stack Bowls Three & 0.0356 & Medium & M--L--H--L--H--M--L & L+M+H & 3 \\
Stack Bowls Two & 0.0328 & Medium & M--L--H--M--L & L+M+H & 3 \\
Stamp Seal & 0.0343 & Medium & H--L & L+H & 2 \\
Turn Switch & 0.0320 & Medium & M & M & 1 \\

\end{longtable}
\endgroup

\section{Additional Experimental Results in Real World}

\textbf{Training and Evaluation Details.}
We evaluate \modelname{} on three representative tasks from the RoboTwin 2.0 benchmark:

(1) Place Object Basket: the robot must grasp one of five kinds of objects and place it in the basket; success is recorded when the object is put in the basket.

(2) Stack Blocks Two: the robot must grasp two blocks and stack one on the other one; success is recorded when the block is centrally stacked on the other one.

(3) Stack Bowls Two: the robot must grasp two bowls and stack one on the other one; success is recorded when the bowl is centrally stacked on the other one.

Each task includes 50 demonstrations used for training. Demonstrations are collected through keyboard-based teleoperation. 
In the hard setting, distractor layouts are randomly sampled across episodes but kept identical across compared models. Success is judged by a blinded human evaluator according to the task-specific criteria above. 

\begin{figure*}[!htb]
  \centering
   \includegraphics[width=\linewidth]{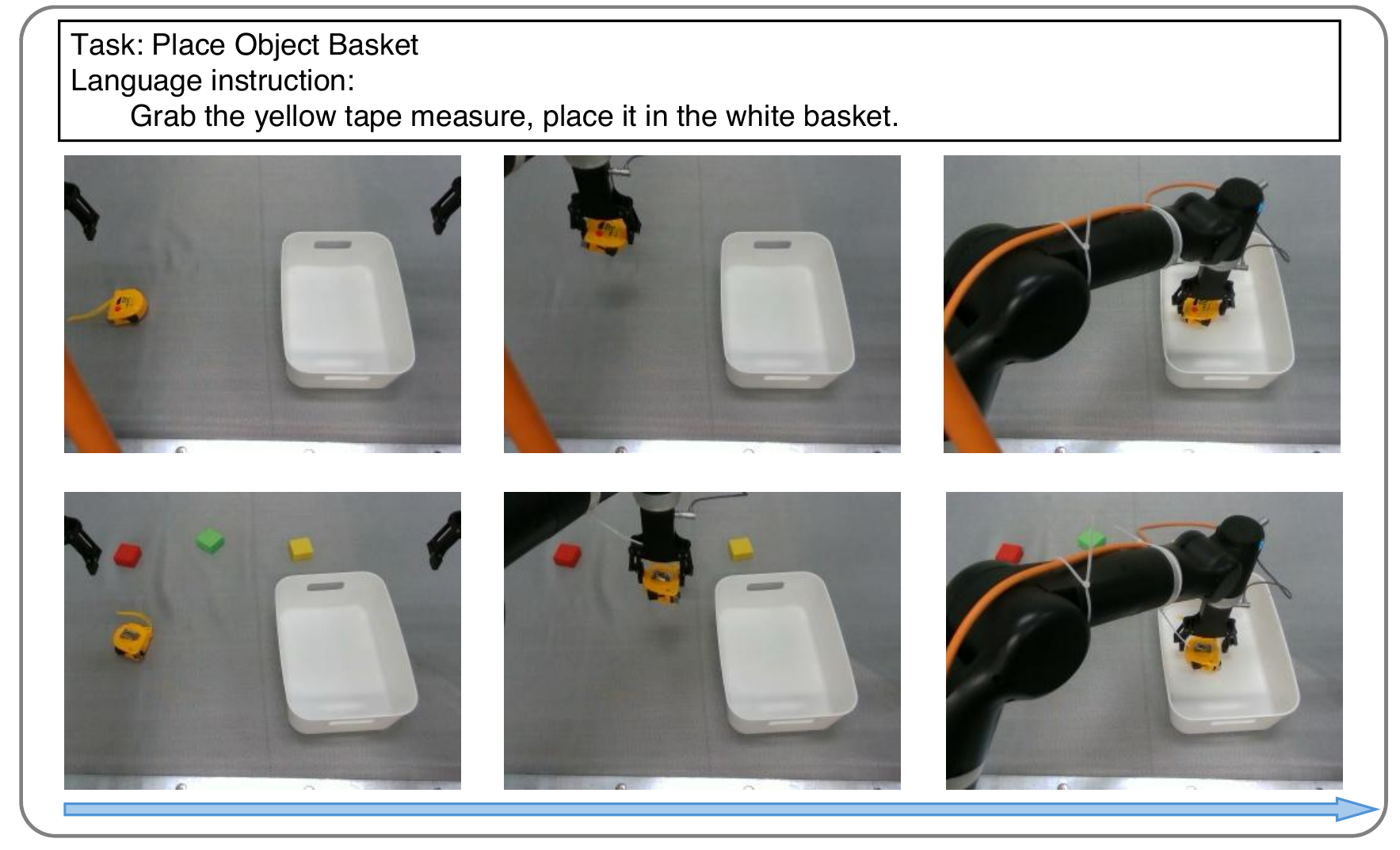}
   \caption{Visualization of Place Object Basket task execution with $\pi_{0.5}$+ActionUNet. The top row shows the keyframes of the task execution in the easy setting, while the bottom row presents the keyframes in the hard setting.}
   \label{fig: place_object_basket}
\end{figure*}

\begin{figure*}[!htb]
  \centering
   \includegraphics[width=\linewidth]{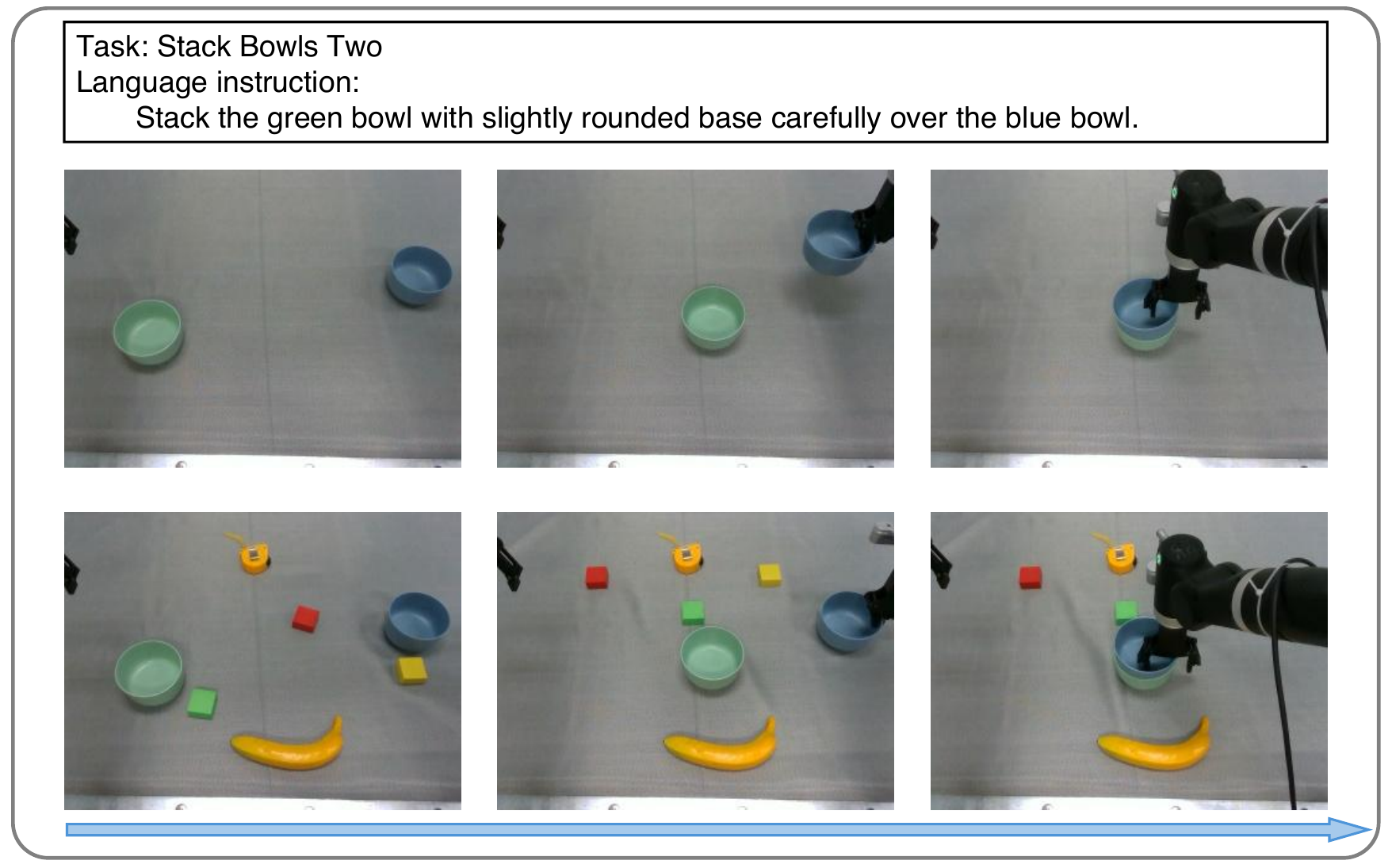}
   \caption{Visualization of Stack Bowls Two task execution with $\pi_{0.5}$+ActionUNet. The top row shows the keyframes of the task execution in the easy setting, while the bottom row presents the keyframes in the hard setting.}
   \label{fig: stack_bowls_two}
\end{figure*}

\begin{figure*}[!htb]
  \centering
   \includegraphics[width=\linewidth]{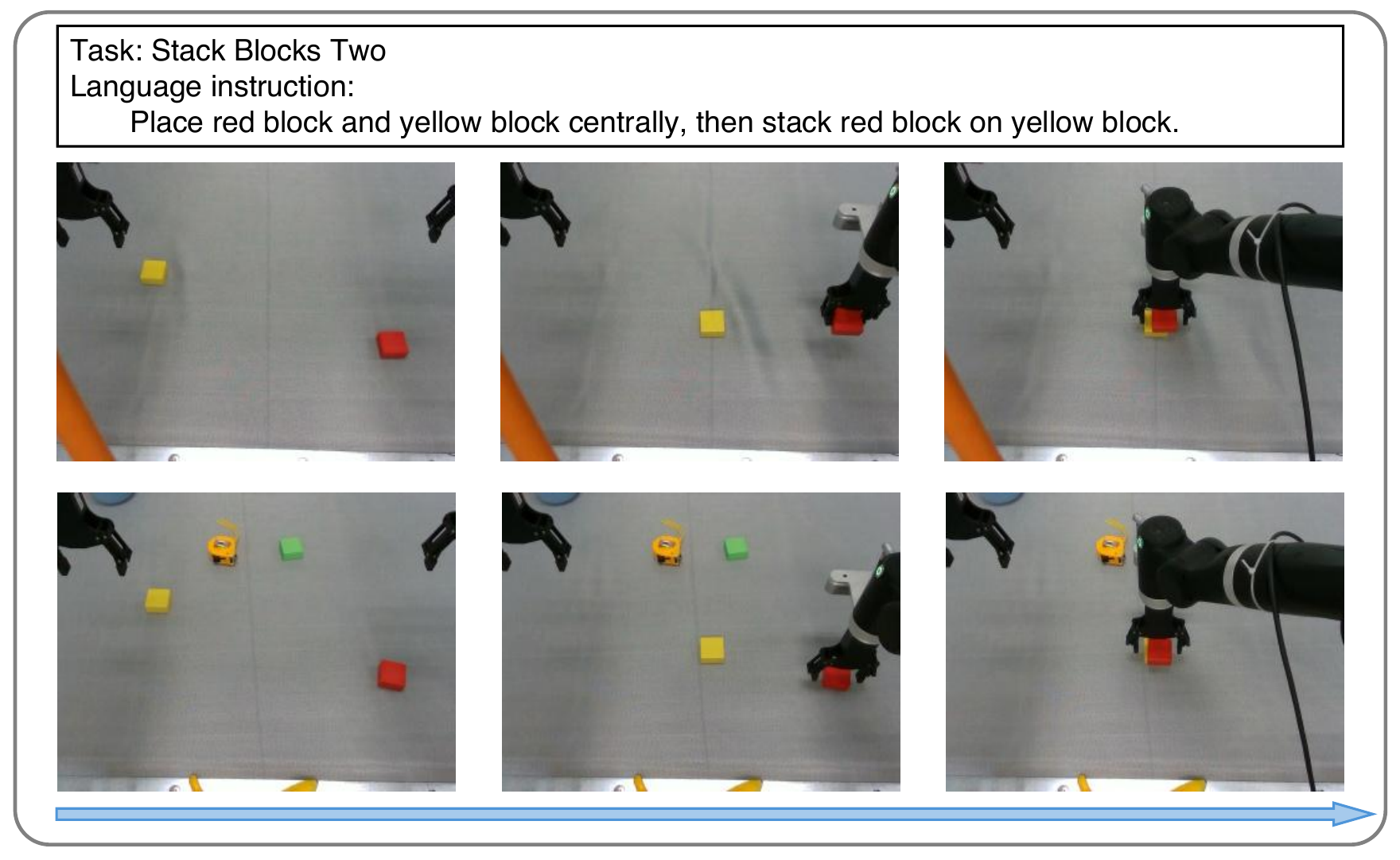}
   \caption{Visualization of Stack Blocks Two task execution with $\pi_{0.5}$+ActionUNet. The top row shows the keyframes of the task execution in the easy setting, while the bottom row presents the keyframes in the hard setting.}
   \label{fig: stack_blocks_two}
\end{figure*}

\textbf{Visualizations.}
We further provide visualizations of real-world task executions in Fig.~\ref{fig: place_object_basket}, Fig.~\ref{fig: stack_bowls_two}, and Fig.~\ref{fig: stack_blocks_two}. Across Place Object Basket, Stack Bowls Two, and Stack Blocks Two, $\pi_{0.5}$+ActionUNet successfully completes the manipulation process in both easy and hard settings, demonstrating stable execution under additional distractor objects.

\section{Architecture Exploration: Where Should Multi-scale Features Enter the Continuous Decoder?}

We further study an alternative design that directly injects multiple U-Net decoder features into different layers of the SIREN decoder. Specifically, given decoder features from the finest, middle, and coarsest resolutions, this variant conditions the first, second, and third SIREN layers on these features, respectively. This design appears natural because the temporal U-Net produces a feature pyramid whose levels correspond to different temporal resolutions. However, as shown in Table~\ref{tab:arch_ablation}, this layer-wise pyramid conditioning does not improve performance and is consistently worse than our default ActionUNet design.

We attribute this result to a mismatch between the temporal scale hierarchy and the SIREN layer hierarchy. The depth of a SIREN represents nonlinear composition in an implicit continuous field, but it does not provide an ordered coarse-to-fine temporal scale axis. Injecting U-Net features with different temporal resolutions into different sinusoidal layers, therefore forces the continuous decoder to simultaneously align multi-scale features and decode smooth trajectories. Because sine activations are sensitive to feature-induced phase shifts, unaligned coarse and fine conditions can interfere inside the hidden field, producing less stable action predictions. In contrast, ActionUNet first reconciles multi-scale information in the temporal action feature space through the U-Net decoder and then passes a single temporal-aligned refined feature to the conditional SIREN. This separation lets the U-Net model have a hierarchical temporal structure, while the SIREN focuses on continuous and smooth action decoding, which explains the stronger performance of our default design.

\begin{table}[!t]
  \centering
  \caption{Architecture ablation on where multi-scale features enter the continuous decoder. 
  Layer-wise Pyramid Conditioning directly injects U-Net decoder features at different temporal resolutions into different SIREN layers. 
  All non-baseline variants are built on $\pi_{0.5}$.}
  
  \footnotesize
\renewcommand{\arraystretch}{1.12}
\setlength{\tabcolsep}{8pt}
    \begin{tabular}{l*{2}{c}|*{2}{c}|*{2}{c}}
      \toprule
      \textbf{Task}
      & \multicolumn{2}{c|}{\textbf{$\pi_{0.5}$}}
      & \multicolumn{2}{c|}{\makecell[c]{\textbf{+ Layer-wise Pyramid}\\\textbf{Condition}}}
      & \multicolumn{2}{c}{\textbf{+ \modelname{}}} \\
      & \textbf{Easy} & \textbf{Hard}
      & \textbf{Easy} & \textbf{Hard}
      & \textbf{Easy} & \textbf{Hard} \\
      \midrule

      Move Can Pot
      & 66 & 52
      & 71 & 64
      & \textbf{77} & \textbf{69} \\

      Stack Blocks Two
      & 68 & 24
      & \textbf{74} & 25
      & \textbf{74} & \textbf{35} \\

      Beat Block Hammer
      & 77 & 28
      & 86 & 34
      & \textbf{89} & \textbf{43} \\

      \midrule
      \rowcolor{gray!8}
      \textbf{Average}
      & 70.3 & 34.7
      & 77.0 & 41.0
      & \textbf{80.0} & \textbf{49.0} \\

      \textbf{Gain over $\pi_{0.5}$}
      & -- & --
      & +6.7 & +6.3
      & \textbf{+9.7} & \textbf{+14.3} \\
      \bottomrule
    \end{tabular}
  
  \vspace{-1.5mm}
  \label{tab:arch_ablation}
\end{table}
\section[Ablation on the Mat\'ern Length-Scale Parameter rho]{Ablation on the Mat\'ern Length-Scale Parameter $\rho$}
\label{app:ablation-rho}

We evaluate the length-scale parameter $\rho$ in the Mat\'ern
aggregation kernel. The parameter $\rho$ controls temporal bandwidth of
the fixed Mat\'ern weights which aggregate neighboring local SIREN fields.

\begin{figure}[t]
    \centering
    \includegraphics[width=0.6\linewidth]{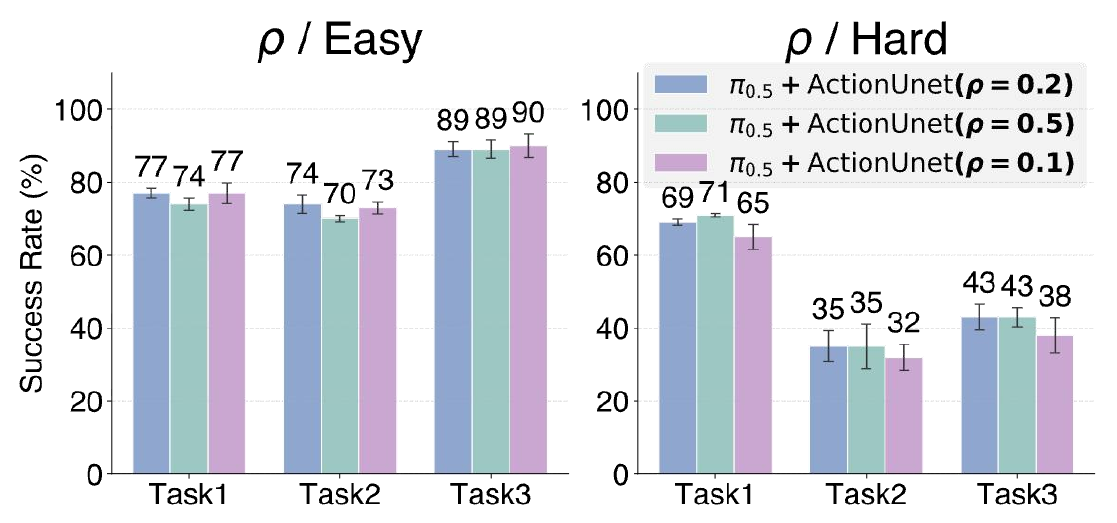}
    \caption{
    Ablation study on the Mat\'ern length-scale parameter $\rho$.
    Changing $\rho$ within the tested range only introduces minor performance
    variations, showing that ActionUNet is robust to the choice of the
    Mat\'ern aggregation bandwidth.
    }
    \label{fig:ablation-rho}
\end{figure}

As shown in Fig.~\ref{fig:ablation-rho}, the Mat\'ern length-scale parameter $\rho$ has only a minor effect: the frozen Mat\'ern kernel mainly provides lightweight aggregation, while the conditional SIREN does the heavy lifting in representing the continuous action field.
We thus fix $\rho=0.2$ in all experiments.

\section{Quantitative Analysis of Trajectory Accuracy and Smoothness}
\label{sec:trajectory_quantitative}

To quantitatively support the trajectory analysis in
Fig.~\ref{fig:vis_result}, we compute two metrics on four RoboTwin 2.0 tasks:
Move Can Pot, Stack Blocks Two, Beat Block Hammer, and Pick Dual Bottles.

\textbf{Jerk RMS} is the root mean square of the third temporal derivative of
the end-effector position averaged over the action horizon. Lower values
indicate smoother trajectories with less high-frequency jitter.
\textbf{Manipulation Drift} is the positional offset of the end-effector at the
first grasp attempt relative to the target object. It measures spatial accuracy
in the critical grasping phase, with lower values indicating more precise
localization.

\begin{table}[!t]
  \centering
  \caption{
  Quantitative trajectory analysis on four RoboTwin 2.0 tasks.
  Lower Manipulation Drift and Jerk RMS indicate better spatial accuracy and
  smoother trajectories, respectively.}
  \label{tab:trajectory_metrics}

  {\footnotesize
  \renewcommand{\arraystretch}{1.12}

  \begin{tabular*}{0.78\textwidth}
    {@{\extracolsep{\fill}}llccc@{}}
    \toprule
    \textbf{Setting}
    & \textbf{Method}
    & \textbf{Success Rate (\%) $\uparrow$}
    & \textbf{Drift (mm) $\downarrow$}
    & \textbf{Jerk RMS (m/s$^3$) $\downarrow$} \\
    \midrule

    \multirow{4}{*}{Easy}
    & $\pi_{0.5}$
    & 69.0 & 32.53 & 32.90 \\

    & + U-Net
    & 67.5 & 29.07 & 34.49 \\

    & + SIREN
    & 75.5 & 28.85 & \textbf{29.43} \\

    & + \modelname{}
    & \textbf{77.8} & \textbf{27.60} & 32.70 \\

    \midrule

    \multirow{4}{*}{Hard}
    & $\pi_{0.5}$
    & 34.5 & 81.64 & 48.31 \\

    & + U-Net
    & 37.3 & 76.05 & 49.34 \\

    & + SIREN
    & 33.3 & 83.03 & \textbf{42.01} \\

    & + \modelname{}
    & \textbf{46.5} & \textbf{62.28} & 45.88 \\

    \bottomrule
  \end{tabular*}
  }
\end{table}

The results confirm that the SIREN-only variant achieves the lowest Jerk RMS
and the best smoothness, but its Hard-setting success rate decreases from
34.5\% to 33.3\%, while Manipulation Drift increases from 81.64\,mm to
83.03\,mm. This shows that temporal smoothing alone is insufficient to explain
the robustness gains.

In contrast, the U-Net-only variant improves Hard-setting success from 34.5\%
to 37.3\% and reduces Manipulation Drift from 81.64\,mm to 76.05\,mm, while
increasing Jerk RMS. The full \modelname{} achieves the highest Hard-setting
success rate of 46.5\% and the lowest drift of 62.28\,mm while maintaining
relatively low jerk. These results indicate that the temporal U-Net provides
the main robustness improvement through multi-scale temporal feature
aggregation, while the SIREN decoder complements it by resolving
fusion-induced temporal discontinuities.
\section{Module-Wise Robustness Ablation on LIBERO-Plus}
\label{sec:libero_plus_module_ablation}

To directly separate the contribution of multi-scale temporal refinement from
that of the smoothing decoder, we conduct the module-wise ablation on
LIBERO-Plus. All variants are trained under the same LIBERO protocol and
directly evaluated on the seven controlled LIBERO-Plus perturbations.
We repeat the evaluation three times and report the mean values and standard
errors.

Performance on LIBERO-Plus provides a direct measure of robustness, as the
models are trained on clean data but tested under diverse observation
perturbations.

\begin{table}[!t]
  \centering
  \caption{
  Module-wise ablation on LIBERO-Plus. All variants are trained on LIBERO and
  directly evaluated under seven controlled perturbations. We report mean
  success rate (\%) $\pm$ standard error over three repeated evaluations.}
  \label{tab:libero_plus_module_ablation}

  {\footnotesize
  \renewcommand{\arraystretch}{1.12}
  \setlength{\tabcolsep}{4.5pt}

  \begin{tabular}{@{}lcccccccc@{}}
    \toprule
    \textbf{Method}
    & \textbf{Camera}
    & \textbf{Robot}
    & \textbf{Lang.}
    & \textbf{Light}
    & \textbf{Back.}
    & \textbf{Noise}
    & \textbf{Layout}
    & \textbf{Avg.} \\
    \midrule

    $\pi_{0.5}$
    & 48.4\se{0.65}
    & 48.0\se{1.75}
    & 67.4\se{0.65}
    & 93.0\se{0.20}
    & 87.1\se{0.30}
    & 51.1\se{1.20}
    & 81.8\se{0.50}
    & 66.0\se{0.40} \\

    + U-Net
    & 50.8\se{0.18}
    & 52.4\se{0.51}
    & 71.3\se{0.34}
    & 93.8\se{0.86}
    & 88.7\se{0.86}
    & 52.9\se{0.17}
    & 83.1\se{0.31}
    & 68.4\se{0.18} \\

    + SIREN
    & 47.4\se{0.23}
    & 53.1\se{0.30}
    & 67.4\se{0.25}
    & 92.7\se{0.55}
    & 86.8\se{0.42}
    & 48.7\se{0.63}
    & 81.4\se{0.53}
    & 66.1\se{0.17} \\

    \rowcolor{gray!8}
    + \modelname{}
    & \textbf{56.4}\se{0.49}
    & \textbf{55.2}\se{1.55}
    & \textbf{74.6}\se{0.51}
    & \textbf{94.8}\se{0.06}
    & \textbf{89.7}\se{0.20}
    & \textbf{60.2}\se{0.97}
    & \textbf{85.5}\se{0.56}
    & \textbf{72.0}\se{0.32} \\

    \bottomrule
  \end{tabular}
  }
\end{table}

Under LIBERO-Plus perturbations, U-Net-only improves the average success rate
by 2.4 percentage points and improves all seven perturbation categories. In
contrast, SIREN-only improves the average by only 0.1 points: it improves the
Robot perturbation, matches the baseline under Language, and decreases
performance under the other perturbations.

The U-Net therefore exhibits the broad standalone robustness effect under
observation corruptions. The full \modelname{} achieves the best performance
under every perturbation and improves the average by 6.0 percentage points,
substantially exceeding either individual module.

These results show that the U-Net alone provides more consistent performance
gains across observation perturbations. Together with the RoboTwin 2.0
module-wise analysis, these results support that the multi-scale scheme
improves the robustness of downstream action-feature prediction, rather than
making the VLM encoder itself more robust.
\section{Precision-Demanding Manipulation on RoboDojo}
\label{sec:robodojo_precision}

We evaluate \modelname{} on high-precision manipulation tasks from
RoboDojo~\cite{robodojo}, specifically on the precision dimension, which
contains tasks requiring strict spatial accuracy and contact-rich control.

We evaluate two representative precision tasks: \emph{Insert Tubes}, where the
robot must insert tubes into narrow holes, and \emph{Build Tower}, where the
robot must precisely arrange wooden blocks and boards into a stable tower.
We compare $\pi_{0.5}$ and $\pi_{0.5}$+\modelname{} under the same training
protocol with 100 demonstrations and single-task fine-tuning. Each evaluation
includes 50 episodes, and we repeat it three times to calculate the mean
$\pm$ standard error.

In addition to success rate, we report the RoboDojo Progress Score and Jerk RMS.
Progress Score measures partial completion. Insert Tubes assigns 20, 40, or
100 points when one, two, or three tubes satisfy the position, insertion-depth,
and orientation conditions; Build Tower assigns 10, 30, or 100 points for
completing the bottom structure, middle structure, or full tower.
Jerk RMS is the root mean square of the third temporal derivative of
end-effector position over the full episode. GT Jerk RMS is computed from the
demonstration trajectories.

\begin{table}[!t]
  \centering
  \caption{
  Evaluation on precision-demanding RoboDojo tasks. We report mean success
  rate and Progress Score $\pm$ standard error over three repeated evaluations.
  Lower Jerk RMS indicates smoother trajectories.}
  \label{tab:robodojo_precision}

  {\footnotesize
  \renewcommand{\arraystretch}{1.12}

  \begin{tabular*}{0.86\textwidth}
    {@{\extracolsep{\fill}}llcccc@{}}
    \toprule
    \textbf{Task}
    & \textbf{Method}
    & \textbf{Success (\%) $\uparrow$}
    & \textbf{Progress $\uparrow$}
    & \textbf{Jerk RMS $\downarrow$}
    & \textbf{GT Jerk RMS} \\
    \midrule

    \multirow{2}{*}{Insert Tubes}
    & $\pi_{0.5}$
    & 4.7\se{3.1}
    & 22.0\se{3.2}
    & 127.81
    & 50.62 \\

    & + \modelname{}
    & \textbf{14.0}\se{2.0}
    & \textbf{30.8}\se{0.4}
    & \textbf{122.99}
    & 50.62 \\

    \midrule

    \multirow{2}{*}{Build Tower}
    & $\pi_{0.5}$
    & 26.0\se{2.0}
    & 37.3\se{0.9}
    & 119.75
    & 49.54 \\

    & + \modelname{}
    & \textbf{34.7}\se{2.3}
    & \textbf{42.5}\se{1.5}
    & \textbf{109.53}
    & 49.54 \\

    \bottomrule
  \end{tabular*}
  }
\end{table}

On Insert Tubes, \modelname{} improves the success rate by 9.3 percentage
points and the Progress Score by 8.8 points. On Build Tower, it improves the
success rate by 8.7 percentage points and the Progress Score by 5.2 points.
These consistent gains in both complete success and partial task progress
demonstrate that \modelname{} remains effective on precision-demanding
manipulation tasks.

The quantitative results further show that \modelname{} is only slightly
smoother than the $\pi_{0.5}$ baseline and remains substantially less smooth
than the ground-truth trajectory. On Insert Tubes, Jerk RMS decreases from
127.81 to 122.99\,m/s$^3$, compared with a GT Jerk RMS of 50.62\,m/s$^3$.
On Build Tower, it decreases from 119.75 to 109.53\,m/s$^3$, compared with a
GT Jerk RMS of 49.54\,m/s$^3$. The simultaneous improvements in complete
success, partial task progress, and trajectory smoothness indicate that the
lower Jerk RMS does not come from suppressing the fine-grained motions required
for precise manipulation. These results support that \modelname{} remains
effective on precision-demanding tasks without over-smoothing the trajectory.


\end{document}